\documentclass{article}

\usepackage{microtype}
\usepackage{graphicx}
\usepackage{subfigure}
\usepackage{booktabs} 

\usepackage{hyperref}

\usepackage{color}
\usepackage{helvet}
\usepackage{courier}
\usepackage{comment}
\usepackage{subfigure}
\usepackage{graphicx}
\usepackage{amsmath}
\usepackage{amsthm}
\usepackage{amsmath}
\usepackage{amssymb}
\theoremstyle{plain}
\newtheorem{thm}{Theorem}[section]
\newtheorem{lem}[thm]{Lemma}

\theoremstyle{definition}
\newtheorem{defn}{Definition}[section]

\usepackage[accepted]{icml2018}

\icmltitlerunning{Randomness in Deconvolutional Networks for Visual Representation}

\begin{document}

\twocolumn[
\icmltitle{Randomness in Deconvolutional Networks for Visual Representation}






Kun He$^{1,2}$, Jingbo Wang$^1$, Haochuan Li$^3$, Yao Shu$^1$, Mengxiao Zhang$^3$, Man Zhu$^1$, Liwei Wang$^3$, John E. Hopcroft$^2$  \\

$^1$School of Computer Science, Huazhong University of Science and Technology, Wuhan, China; \\
$^2$Department of Computer Science, Cornell University, Ithaca, USA; \\
$^3$Department of Machine Intelligence, Peking University, Beijing, China. \\
Correspondence to: Kun He: brooklet60@hust.edu.cn, Liwei Wang: wanglw@cis.pku.edu.cn.




\icmlkeywords{Machine Learning, ICML}

\vskip 0.3in
]




\begin{abstract}
Toward a deeper understanding on the inner work of deep neural networks, we investigate CNN (convolutional neural network) using DCN (deconvolutional network) and randomization technique, and gain new insights for the intrinsic property of this network architecture. For the random representations of an untrained CNN, we train the corresponding DCN to reconstruct the input images.  Compared with the image inversion on pre-trained CNN, our training converges faster and the yielding network exhibits higher quality for image reconstruction. It indicates there is rich information encoded in the random features; the pre-trained CNN may discard information irrelevant for classification and encode relevant features in a way favorable for classification but harder for reconstruction. We further explore the property of the overall random CNN-DCN architecture. Surprisingly, images can be inverted with satisfactory quality. Extensive empirical evidence as well as theoretical analysis are provided. 
\end{abstract}

\section{Introduction}
\label{sec:Intro}

Since the first introduction in 1990's~\cite{LeCun1989}, convolutional neural networks (CNNs) have demonstrated impressive performance for various computer vision tasks, 
and an essential element is to understand the deep representations for intermediate layers. 
A variety of visualization techniques have been developed in order to unveil the feature representation and hence the inner mechanism of CNNs~\cite{zeiler2014visualizing,Mahendran2015CVPR,yosinski2015understanding, xu2015show}, however, our understanding of how these deep learning models operate remains limited. 
In this paper we propose to apply randomization, the assignment of random weights, on deconvolutional networks for the visual representation, 
and permit deeper understanding on the intrinsic property of the convolutional architecture. 
Two techniques are closely related to our work, \textit{deconvolution} and \textit{randomization}. 

Deconvolutional networks (DCNs) are commonly used for deep feature visualization. 
In the seminal work of Zeiler et al.~\yrcite{zeiler2014visualizing}, they propose to use a multi-layered deconvolutional network~\cite{Zeiler2011ICCV} to project the feature activations back to the input pixel space, and show that the features have many intuitively desirable properties such as
compositionality, increasing invariance and class discrimination for deeper layers. 
Dosovitskiy et al.~\yrcite{Alexey2016CVPR} design a deconvolution variant which they call the up-convolutional neural network 
to invert image representations learned from a pre-trained CNN, and conclude that features in higher layers preserve colors and rough contours of the images and discard information irrelevant for the classification task that the convolutional model is trained on. As there is no back propagation, their reconstruction is much quicker than the representation inverting method based on gradient descent~\cite{Mahendran2015CVPR}. 

Randomization on neural networks can be tracked back to the 1960's where the bottom-most layer of shallow networks consisted of random binary connections~\cite{Block1962}.  
In recent years, largely motivated by the fact that ``\textit{randomization is computationally cheaper than optimization}", randomization has been resurfacing repeatedly in the machine learning literature~\cite{wang2017Review}. For optimization problems such as regression or classification, this technique is used to stochastically assign a subset of the network weights
to derive a simpler problem~\cite{Igelnik1995, Rahimi2009weighted}.  
Specifically, they compute a weighted sum of the inputs after passing them through a bank of arbitrary randomized nonlinearities,
such that the resulting optimization task is formulated as a linear least-squares problem.
Empirical comparisons as well as theoretical guarantees are provided for the approximation~\cite{Rahimi2008Uniform, Rahimi2009weighted,AroraBGM14}.
Other related works include random kernel approximation~\cite{Rahimi2007NIPS, Sinha2016NIPS}.

Specifically on convolutional neural networks (CNNs), there are a few lines of work considering randomization.
Jarrett et al.~\yrcite{JarrettICCV09} observe that, on a one-layer convolutional pooling architecture, random weights perform only slightly worse than pre-trained weights. 
Andrew et al.~\yrcite{AndrewNg11ICML} prove that certain convolutional pooling architectures with random weights are inherently frequency selective and translation invariant, and argue that these properties underlie their performance. 
He et al.~\yrcite{He2016NIPS} accomplish three popular visualization tasks, image inversion, texture synthesize and style transfer, using random weight CNNs.
Daniely et al.~\yrcite{DanielyNIPS2016} extend the scope from fully-connected and convolutional networks to a broad family of architectures by building a general duality between neural networks and compositional kernel Hilbert spaces and proving that random networks induce representations which approximate the kernel space.

Motivated by the intuition that ``\textit{random net may be theoretically easier to comprehend than the complicated welltrained net}", and that it may reveal the intrinsic property of the network architecture,
we use randomization to explore the convolution followed by deconvolution architecture, and provide theoretical analysis on the empirical observations. 
Our goal is toward a deeper understanding on the inner mechanism of deep convolutional networks. 

Our main contributions are as follows:  

First, we propose a random weight CNN subsequently connecting a trained deconvolutional network(DCN) to reconstruct images. 
By means of CNN-DCN architecture,  either fixing the random weights assigned to CNN model or utilising the pre-trained CNN model, we train the corresponding DCN to invert the input image. 
The DCN architecture uses the inverted layer sequence of the CNN, as in~\cite{Alexey2016CVPR}.
Compared with the inversion on pre-trained CNN, the random approach can train the corresponding DCN model more quickly and invert the image with higher quality. 
The results explicitly reveal that random weight CNN can encode rich feature information of the inputs while pre-trained CNN may discard feature information irrelevant for classification and encode relevant feature in a way favourable for classification but harder for image reconstruction.

Second, we present the overall random CNN-DCN architecture to further investigate the randomness in CNNs, i.e. there is no training at all for inverting the inputs that pass their information through a random weight convolutional network.
Surprisingly, the image is inverted with satisfactory quality. The geometric and photometric feature of the inputs are well preserved.
We argue that this is due to the intrinsic property of the CNN-DCN architecture. 
We provide empirical evidence as well as theoretical analysis on the reconstruction quality, 
and bound the error in terms of the number of random nonlinearities, the network architecture and the distribution of the random weights.

\section{Preliminaries}
\label{sec:Architecture}

\subsection{Deconvolutional network architecture}

For the network architecture, we consider two typical CNNs for the deconvolution, 
VGG16~\cite{simonyan2014VGG} and AlexNet~\cite{krizhevsky2012imagenet}. 
A convolutional layer is usually followed by a pooling layer, except for the last convolutional layer, Conv5. For consistency, we will explore the output after the convolutional layer but before the pooling layer. 
In what follows, ``feature representation'' or ``image representation'' denotes the feature vectors after the linear convolutional operator and the nonlinear activation operator but before the pooling operator for dimension reduction.

\begin{figure}[htbp]
	\centering
	\vspace{-0.5em}
	\includegraphics[width=3.2in]{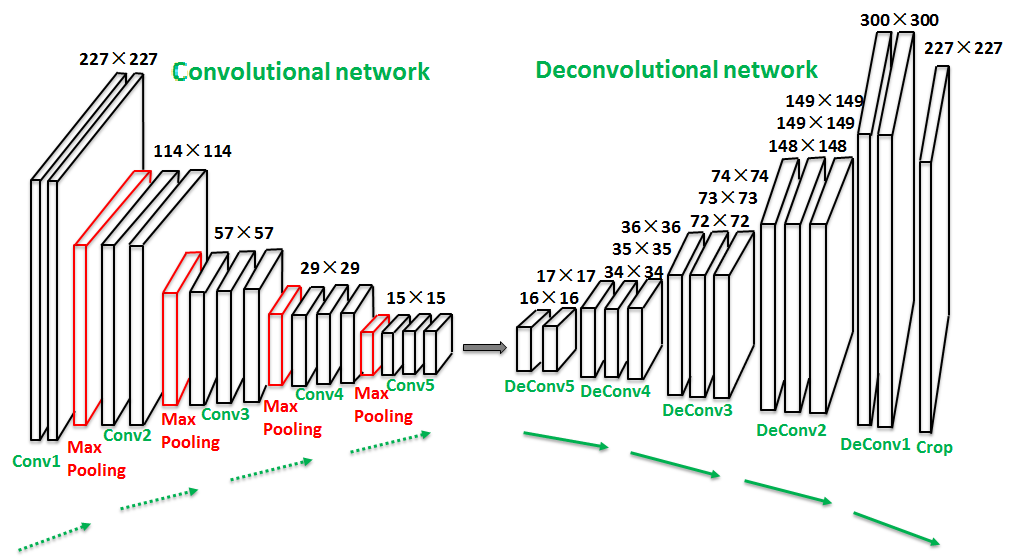}
	\vspace{-0.5em}
	\caption{The CNN-DCN architecture for Conv5 of VGG16.} 
	\label{fig:structure}
	\vspace{-0.5em}
\end{figure}

We build a CNN-DCN architecture on the layer of feature representation to be studied. 
The convolution operator of a deconvolutional layer in DCN is the same as the convolution operator in CNN, and an upsampling operator is applied in DCN to invert the corresponding pooling operator in CNN, as designed in \cite{Alexey2016CVPR}. 
We will focus on the representations of the convolutional layers,
as Dosovitskiy et al. build DCNs for each layer of the pre-trained AlexNet and find that the predicted image from the fully connected layers becomes very vague. 
Figure~\ref{fig:structure} illustrates an example of the VGG16 Conv5-DeConv5 architecture, where Conv5 indicates the sequential layers from Conv1 to Conv5.
For the activation operator, we apply the leaky ReLU nonlinearity with slope 0.2, that is, $r(x) = x$ if $x \geq 0$ and otherwise $r(x) = 0.2x$. At the end of the DCN, a final Crop layer is added to cut the output of DeConv1 to the same shape as the original images.

We build deconvolutional networks on both VGG16 and AlexNet, and most importantly, we focus on the random features of the CNN structure when training the corresponding DCN. Then we do no training for deconvolution and explore the properties of the purely random CNN-DCN architecture.

\subsection{Random distributions}
\label{subsec:Experiment Details}
For the random weights assigned to CNN or DCN, we try several Gaussian distributions with zero mean and various variance
to see if they have different impact on the DCN reconstruction. Subsequent comparison shows 
that a small variance around 0.015 yields minimal inverting loss. 
We also try several other types of random distributions, Uniform, Logistic, Laplace, to study their impact. 
\vspace{-0.5em}
\begin{itemize} 
	\setlength{\itemsep}{0pt}
	\setlength{\parsep}{0pt}
	\setlength{\parskip}{0pt}	
	\item The Uniform distribution is in [-0.04, 0.04), such that the interval equals $[\mu-3\delta, \mu+3\delta]$ where $\mu=0$ and $\delta=0.015$ are parameters for Gaussian distribution. 
	\item The Logistic distribution is 0-mean and 0.015-scale of decay. It resembles the normal distribution in shape but has heavier tails. 
	\item The Laplace distribution is with 0 mean and $2*\lambda^2$ variance ($\lambda = 0.015$), which puts more probability density at 0 to encourage sparsity.
\end{itemize}

\section{Random-CNN-trained-DCN}
\label{sec:ExperimentRanTra}

\subsection{Training method for DCN}

For each intermediate layer, using the feature vectors of all training images, we train the corresponding DCN such that the summation of $L_2$-norm loss between the inputs and the outputs is minimized. Let $\Phi(x_i,w)$ represent the output image of the DCN, in which $x_i$ is the input of the $i^{th}$ image and $w$ is the weights of the DCN. We train the DCN to get the desired weights $w^*$ that minimize the loss. Then for a feature vector of a certain layer, the corresponding DCN can predict an estimation of the \textit{expected pre-image}, the average of all natural images which would have produced the given feature vector. 
\begin{equation}
w^*=\arg\min_{w}L=\arg\min_w \sum\limits_{i}(\Phi(x_i,w)-x_i)^2
\end{equation}
\vspace{-0.5em}

Specifically, we initialize the DCN by the ``MSRA'' method~\cite{he2015delving} based on a modified Caffe~\cite{jia2014caffe,Alexey2016CVPR}. We use the training set of ImageNet~\cite{deng2009imagenet} and the Adam~\cite{kingma2014adam} optimizer with $\beta_1=0.9$, $\beta_2=0.999$ with mini-batch size 32. The initial learning rate is set to 0.0001 and the learning rate gradually decays by the ``multistep'' training. The weight decay is set to 0.0004 to avoid overfitting. 
The maximum number of iterations is set at 200,000 empirically. 

\subsection{Results for inverting the representation}

\begin{figure*}[htbp]
	\centering
	\subfigure[VGG vs. Alexnet]{
		\includegraphics[width=2in]{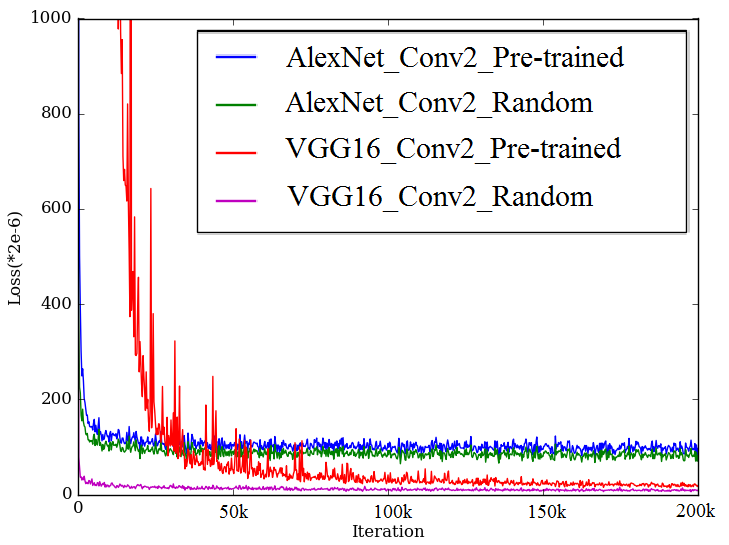}
	}	
	\subfigure[rwVGG]{
		\includegraphics[width=2in]{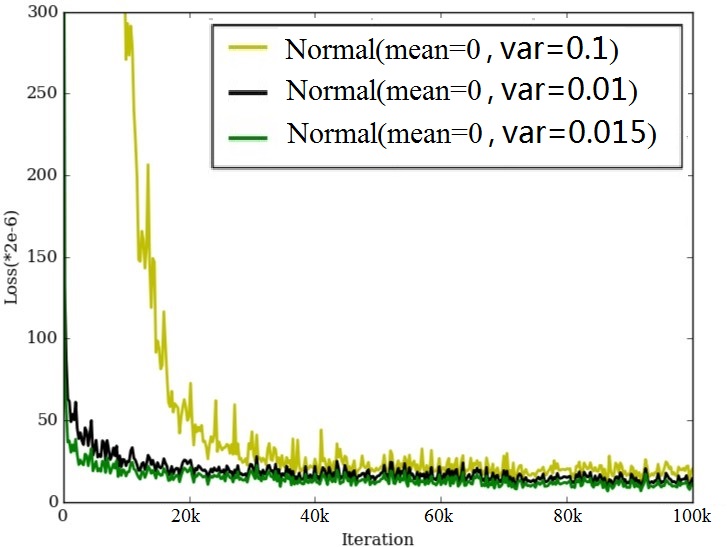}
	}
	\subfigure[rwVGG]{
		\label{fig:other-random-results:b} 
		\includegraphics[width=2in]{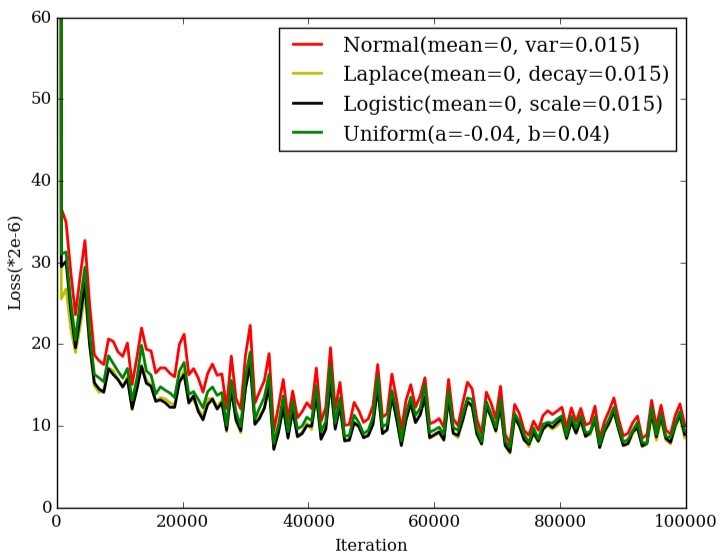}			
	}
	\vspace{-1em}
	\caption{Training loss for the Conv2-DeConv2 architecture. } 
	\label{fig:Alex_vgg_conv2_loss} 
	\vskip -0.2in
	\vskip 0.3in
	\centering
	\vspace{-0.5em}
	\subfigure[Random VGG vs. Pre-trained VGG]{ 
		\includegraphics[height=1.5in]{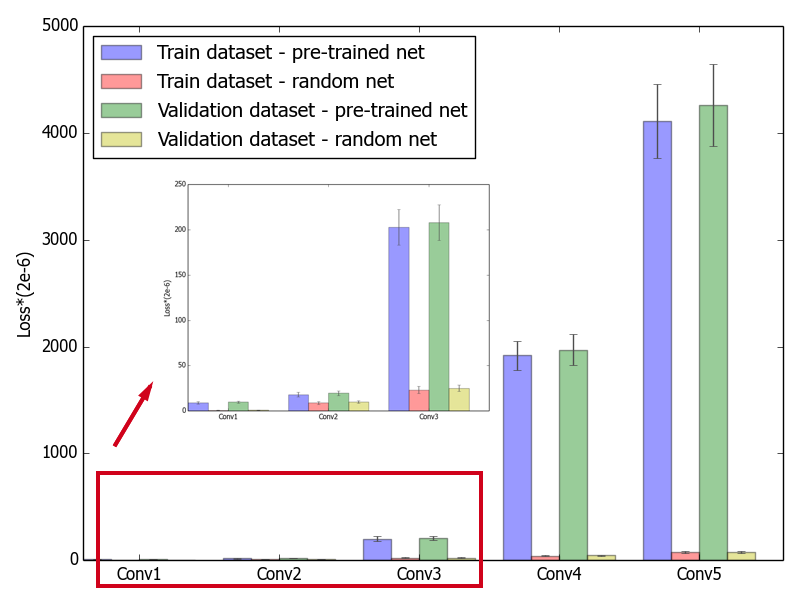} }
	\hspace{0.2in}
	\subfigure[VGG vs. Alexnet on Conv3]{
		\includegraphics[height=1.5in]{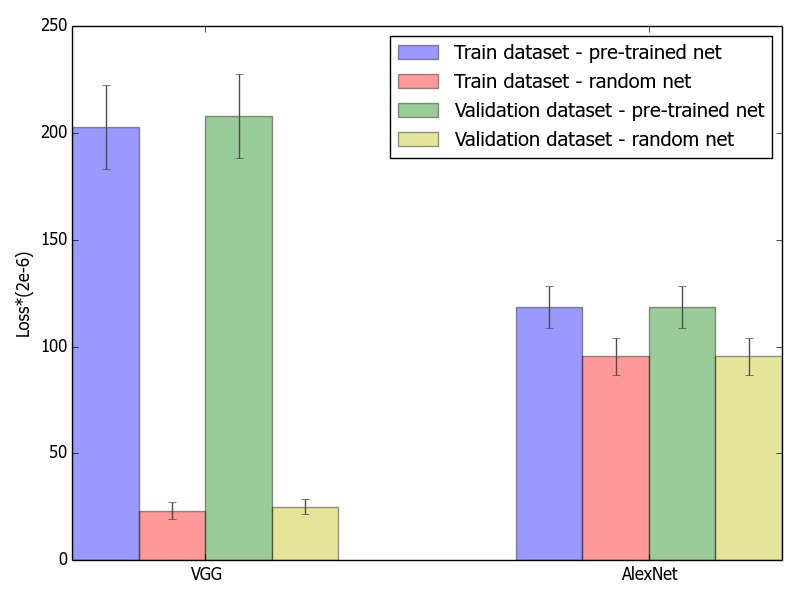} }
	\vspace{-1em}	
	\caption{Comparison on the generalization error.}
	\label{fig:trainVSval}  
	\vskip -0.2in
	\vskip 0.3in
	\centering
	\subfigure[rwVGG]{
		\begin{minipage}[b]{0.45\textwidth}
			\scriptsize{~~~~Original~~~~~~~~~~~~Conv1~~~~~~~~~~~~~~Conv2~~~~~~~~~~~~~~Conv3~~~~~~~~~~~~~~Conv4~~~~~~~~~~~~~~Conv5}
			\label{fig:vgg_alex_cat_reconstruction:a} 
			\includegraphics[height=1.6in]{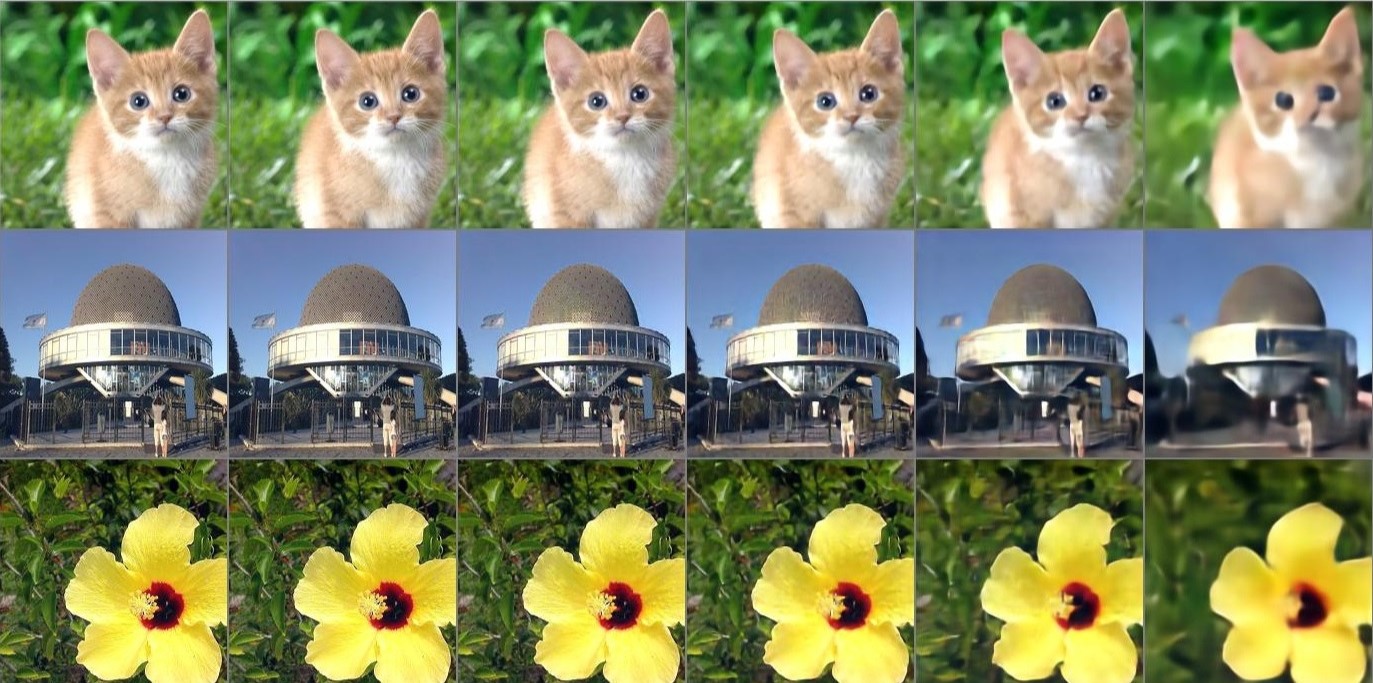} 
	\end{minipage}}
	\hspace{0.4in}
	\subfigure[rwAlexNet]{
		\begin{minipage}[b]{0.4\textwidth}
			\scriptsize{~~~~~Conv1~~~~~~~~~~~~Conv2~~~~~~~~~~~~~Conv3~~~~~~~~~~~~Conv4~~~~~~~~~~~~Conv5}
			\label{fig:vgg_alex_cat_reconstruction:b}  
			\includegraphics[height=1.6in]{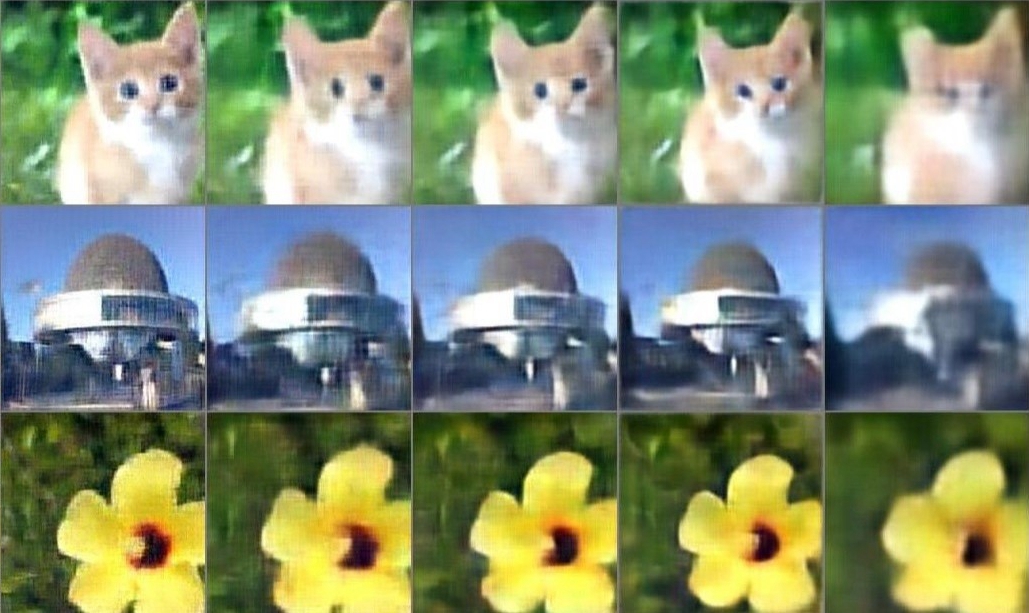}
	\end{minipage}}
	\vspace{-1em}
	\caption{(\textbf{Zoom in for details.}) Reconstructions for representations of different convolutional layers of rwVGG and rwAlexNet. }
	\label{fig:vgg_alex_image_reconstruction} 
	\vspace{-0.5em}
	\vskip 0.2in
	\centerline{\includegraphics[width=16cm]{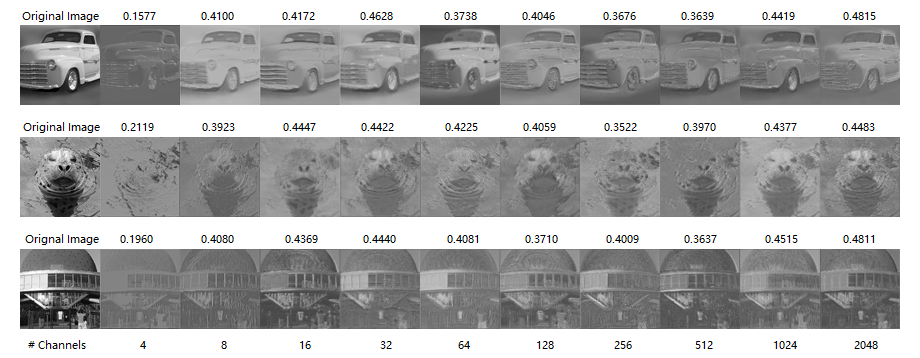}}
	\caption{(\textbf{Zoom in for details.}) Reconstructions on rrVGG Conv1-DeConv1 networks.}
	\label{Conv1_Example}
	\vskip -0.2in	
\end{figure*}

\textbf{Training.}
We observe similar results for the training loss in different layers. 
Take the Conv2-DeConv2 architecture for elaboration, the loss curves during the training process are shown in Figure~\ref{fig:Alex_vgg_conv2_loss}.
Figure~\ref{fig:Alex_vgg_conv2_loss}(a) compares VGG and AlexNet on random as well as pre-trained weights. 
The training for reconstruction converges much quicker on random CNN and yields slightly lower loss, and this trend is more apparent on VGG.
It indicates that by pre-training for classification, CNN encodes relevant features of the input image in a way favorable for classification but harder for reconstruction. 
Also, VGG yields much lower inverting loss as compared with AlexNet. 
Figure~\ref{fig:Alex_vgg_conv2_loss}(b) shows that random filters of different Gaussian distributions on CNN affect the initial training loss, but the loss eventually converges to the same magnitude. 
Figure~\ref{fig:Alex_vgg_conv2_loss}(c) shows that the four different random distributions with appropriate parameters acquire similar reconstruction loss.

\textbf{Generalization.} 
We take 5000 samples from the training set and validation set respectively from ImageNet, and compare their average reconstruction loss.
The statistics is as shown in Figure \ref{fig:trainVSval}, where Conv$k$ represents a Conv$k$-DeConv$k$ architecture.  
Figure \ref{fig:trainVSval}(a) shows that the VGG architecture is good in generalization for the reconstruction,
and random VGG yields much less loss than pre-trained VGG.  
For representations of deeper layers, the inverting loss increases significantly for pre-trained VGG but grows slowly for random VGG.
This means that in deeper layers, the pre-trained VGG discards much more information that is not crucial for classification, leading to a better classifier but a harder reconstruction task. 
Figure \ref{fig:trainVSval}(b) compares VGG and AlexNet on the Conv3-DeConv3 architecture. It shows that Alexnet is also good in generalization for the reconstruction, and the difference between the random and the pre-trained is very small.

\textbf{Reconstruction.} 
Figure~\ref{fig:vgg_alex_image_reconstruction} shows reconstructions from various layers of random VGG and random AlexNet, denoted by rwVGG and rwAlexNet respectively.~\footnote{For input images, the cat is an example image from caffe, and the other two are from the validation set of ImageNet.}  
On both rwVGG and rwAlexNet, the reconstruction quality decays for representations of deeper layers.
The rwVGG structure yields more accurate reconstruction, even on Conv5, which involves 26 convolution operations and 4 max pooling operations.

\begin{figure}[htbp]
	\vspace{-0.5em}	
	\centering
	\vspace{0em}
	\includegraphics[height=1.5in]{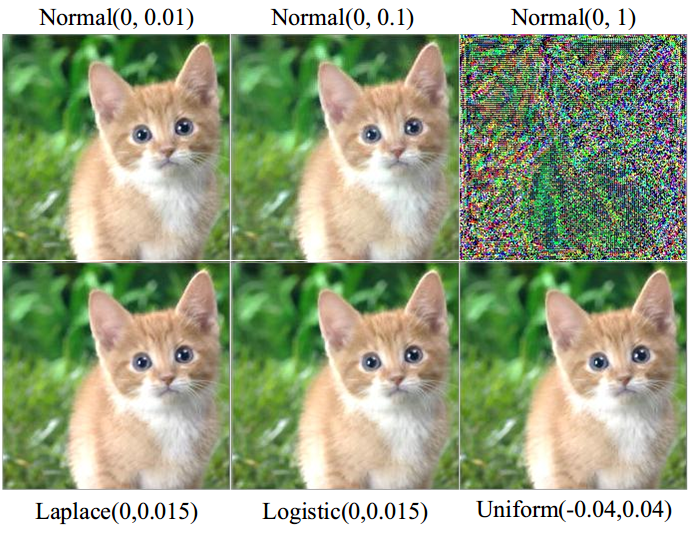}
	\vspace{-1em}	
	\caption{(\textbf{Zoom in for details.}) Reconstructions for representations of rwVGG Conv2 in various random distributions. }
	\label{fig:cat-random-results} 
	\vspace{-1em} 
\end{figure}

Figure~\ref{fig:cat-random-results} shows reconstructions from a cat example image for various distributions of rwVGG Conv2-DeConv2. 
Except for $N(0,1)$, the reconstruction quality is indistinguishable by naked eyes.
It shows that different random distributions work well when we set the random weights relatively sparse.

In a nutshell, it is interesting that random CNN can speed up the training process of the DCN on both VGG and AlexNet, obtain higher quality reconstruction and generalize well for other inputs. 
Regarding weights in the convolutional part as a feature encoding of the original image, then the deconvolutional part can decode from the feature representations encoded by various methods. 
The fact that the random encoding of CNN is easier to be decoded indicates that the training for classification moves the image features of different categories into different manifolds that are moving further apart. Also, it may discard information irrelevant for the classification. The pre-trained CNN benefits the classification but is adverse to the reconstruction.

\section{Random-CNN-random-DCN}
\label{sec:ExperimentRanRan}

In this section we mainly focus on VGG16, and explore the reconstructions on purely random VGG CNN-DCN architecture (denoted by rrVGG for brevity). 
Surprisingly, we find that even the overall CNN-DCN network is randomly connected, the images can be inverted with satisfactory quality.
In other words, the CNN randomly extracts the image features and passes them to the DCN, then in an unsupervised manner the DCN can reconstruct the input image by random feature extraction again. 
Our surprising results show that the random-CNN-random-DCN architecture substantially contributes to the geometric and photometric invariance for the image inversion. 
This reminds us the immortal AI Koan~\cite{Rahimi2009weighted}: 
Sussman said,``I am training a randomly wired neural net to play tic-tac-toe," ``The net is wired randomly as I do not want it to have any preconceptions of how to play." Minsky said, ``I will close my eyes so the room will be empty." And Sussman was enlightened.

In the following, we will systematically explore the reconstruction ability of the rrVGG architecture. For convenience analysis, we use the normal ReLU 
nonlinearity (i.e. \(ReLU(\mathbf{x})=\max(\mathbf{x}, 0)\)), and the dimensions of the convolutional layers of VGG are summarized in Table \ref{table:VGGdimension}. 

For evaluation, we use the Pearson correlation coefficient, $corr \in [-1, 1]$, between the output image and the original image to approximately evaluate the reconstruction quality. 
An extreme $corr$, either positive or negative, indicates a high correlation with the geometric information.
For statistics on the trend, we use the structural similarity (SSIM) index~\cite{Wang2004SSIM}, 
which is more accurate by considering the correlation and dependency of local spatially close pixels,  and more consistent to the perceptron of human eyes.
To remove the discrepancy on colors, we transform the inputs and outputs in grey-scale, 
and in case of negative SSIM value, we invert the luminosity of the grayscale image for calculation, so the value is in $[0, 1]$.

\subsection{Network width versus network depth} 
\begin{figure}[htbp]
	\vspace{-0.5em}	
	\subfigure{
		\begin{minipage}[b]{0.01\textwidth}
			\rotatebox{90}{~~~\scriptsize{Original}~~~~} \\
			\rotatebox{90}{~~~~\scriptsize{Simplified}~~~~}  								
	\end{minipage}}	
	\centering
	\subfigure{
		\begin{minipage}[b]{0.45\textwidth}
			\scriptsize{~~~~~~~Conv1~~~~~~~~~~~~~~Conv2~~~~~~~~~~~~~~~Conv3~~~~~~~~~~~~~Conv4~~~~~~~~~~~~~~Conv5~~}\\
			\vspace{0.05em} 
			\includegraphics[height=1.2in]{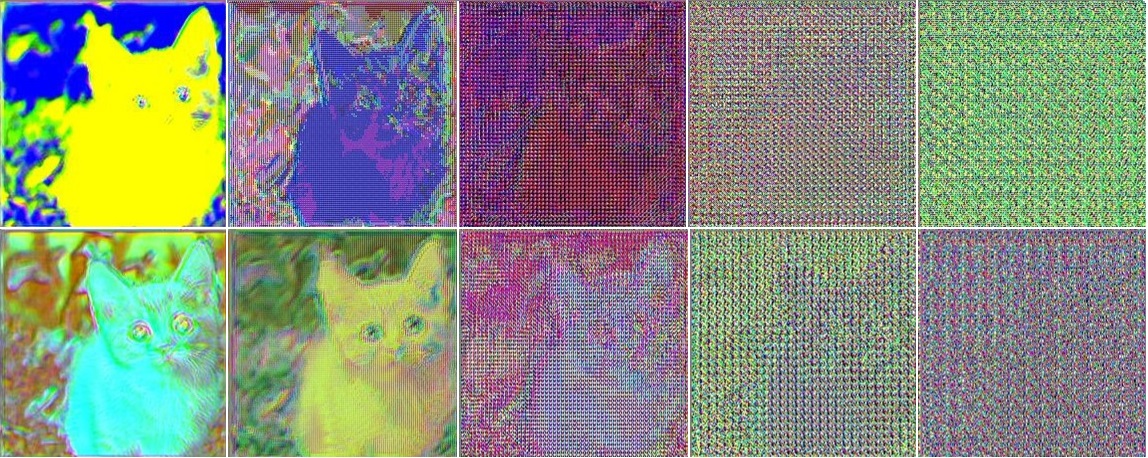}
	\end{minipage}}
	\vspace{-0.5em}
	\caption{(\textbf{Zoom in for details.}) Reconstruction images for the rrVGG architecture, and the simplified rrVGG architecture. All random weights are in $N(0, 0.1)$ distribution. }
	\label{fig:cat_random_random_normal01}
	\vspace{-0.5em} 
\end{figure}

Taking the cat image as an example, we first study the reconstruction quality for different convolutional layers, as shown in the first row of Figure~\ref{fig:cat_random_random_normal01}. The weights are random in $N(0, 0.1)$ and Conv$k$ indicates a Conv$k$-DeConv$k$ architecture. The deeper the random representations are, the coarser the reconstructed images are. Even though there is no training at all, DCN can still perceive geometric positions and contours for Conv1 to Conv3. We can still perceive a very rough contour for the purely random Conv4-DeConv4 architecture, which is already 10 layers deep (Zoom in to see the contours).

In the second row of Figure~\ref{fig:cat_random_random_normal01}, we build a simplified input-Conv$k$-DeCon$k$-output architecture by connecting the data layer directly to Conv$k$ followed by DeConv$k$, then to the output layer. 
And we use the same $N(0, 0.1)$ distribution for the weights. 
The reconstruction quality is similar to that of the first row, indicating that the dimension (width) of the convolutional layer contributes significantly to the reconstruction quality, while the depth of the CNN-DCN network contributes only a little. 

\begin{table}[htbp]
	\centering
	\scalebox{0.8}{
		\begin{tabular}{c|c|c|r}
			\hline
			\hline
			Layer & \# Channels & Shape of feature maps & Dimension\\
			\hline
			\hline
			Data & 3 & 227 $\times$ 227 &  154,587\\
			\hline
			Conv1 & 64 & 227 $\times$ 227 &  3,297,856\\
			\hline
			Conv2 & 128 & 114 $\times$ 114 & 1,663,488 \\
			\hline
			Conv3 & 256 & 57 $\times$ 57 & 831,744\\
			\hline
			Conv4 & 512 & 29 $\times$ 29 & 430,592\\
			\hline
			Conv5 & 512 & 15 $\times$ 15 & 115,200 \\
			\hline
			\hline
		\end{tabular}
	}
	\caption{The dimension of each convolutional layer of VGG.}
	\label{table:VGGdimension}
	\vspace{-0.5em}
\end{table}

As the reconstruction quality decays quickly for deep layers, we argue that it may due to the shape reduction of the feature maps. As shown in Table~\ref{table:VGGdimension}, 
the shape of feature maps, while going through convolutional layers, will be reduced by 1/4 except for the input data layer. 
The convolutional layer will project information of the previous layer to a 1/4 scale shape. So the representations encoded in feature maps will be compressed randomly and it is hard for a random DCN to extract these feature representations. 
However in the previous section, we see that the trained DCN can extract these random deep representations easily and well reconstruct the images. It indicates that almost all information is transformed to the next layer in the random CNN. Without training on DCN, however, the reconstruction through low dimension layers will certainly lose some information.   

\subsection{Results on number of random channels}  

We further explore the reconstruction quality on the number of random channels using the rrVGG Conv1-DeConv1 architecture,
which contains two convolutional operators, one pooling operator, one upsampling operator and another two convolutional operators.
For simplicity, for each network instance we use the same number of channels in all layers except the output layer.
We vary the number of random channels from 4, 8 up to 2048, 
and for each number of channels, we generate 30 rrVGG Conv1-DeConv1 networks and all random weights are in $N(0,0.1)$ distribution.
For input images we randomly pick 50 samples from the ImageNet validation set.

\begin{figure}[htbp]
	\centering
	\subfigure[original rrVGG]{
		\includegraphics[width=1.52in]{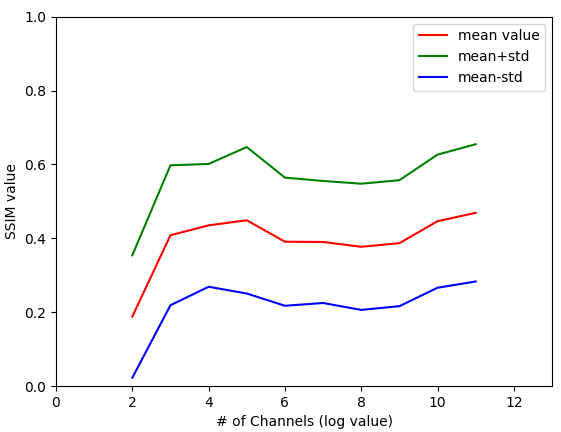}
	}	
	\subfigure[variant rrVGG]{
		\includegraphics[width=1.52in]{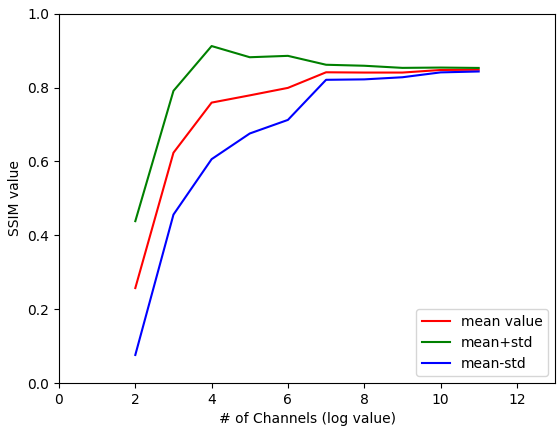}
	}
	\vspace{-1em}
	\caption{Statistics on SSIM for rrVGG Conv1-DeConv1 architecture (mean, mean $\pm$ std). All weights are in N(0,0.1) distribution.}
	\label{SSIM_Conv}
	\vskip -0.0in
	\centering
	\subfigure[original rrVGG]{
		\includegraphics[width=1.52in]{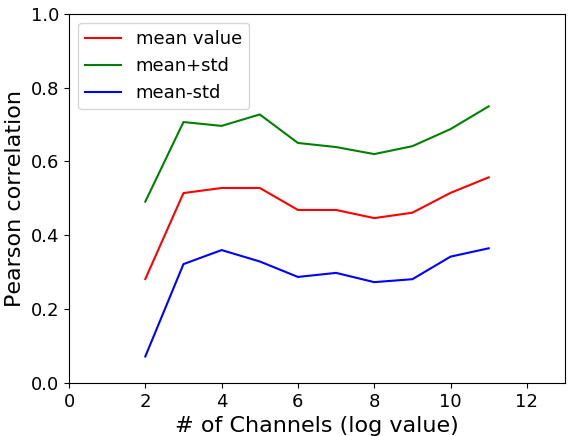}
	}	
	\subfigure[variant rrVGG]{
		\includegraphics[width=1.52in]{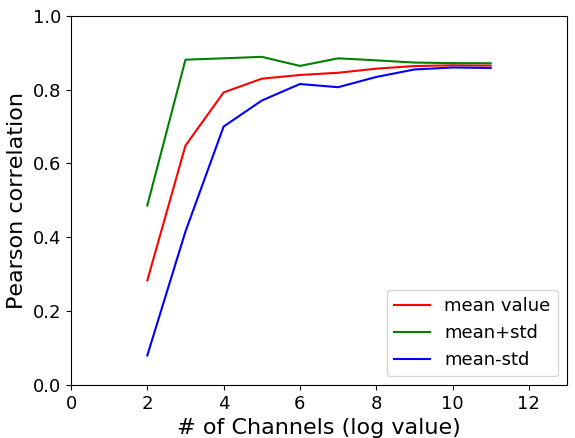}
	}
	\vspace{-1em}
	\caption{Statistics on Pearson correlation for rrVGG Conv1-DeConv1 architecture (mean, mean $\pm$ std). All weights are in N(0,0.1) distribution.}
	\label{Corr_Conv}
\end{figure}

To reduce occasionality on the reconstruction, we transform the inputs and outputs in grey-scale and calculate the average SSIM value and Pearson correlation coefficient on each network, 
then we do statistics (mean and standard deviation) on the 30 average values. 
Figure~\ref{SSIM_Conv} and Figure~\ref{Corr_Conv} show the trends when the number of channels increases, (a) is for the original rrVGG network and (b) is for the variant of rrVGG network. The variant of rrVGG network is almost the same as the original network except that the last convolutional layer is replaced by an average layer, which calculates the average over all the channels of the feature maps next to the last layer. 
We see that the increasing number of random channels promotes the reconstruction quality. 
Similar in spirit to the random forest method, different channels randomly and independently extract some feature from the previous layer, and they are complementary to each other. With a plenty number of random channels we may encode and transform all information to the next layer. The increasing trend and the convergence on the variant of the rrVGG network are much more apparent. We will provide theoretical analysis in the next section.

In Figure~\ref{Conv1_Example}, we pick some input images, and show their output images closest to the mean SSIM value for various number of channels. The SSIM value is on the top of each output image. As the color information is discarded for random reconstruction, we transform the images in grey-scale to show their intrinsic similarity. The increasing number of channels promotes the random reconstruction quality.

\subsection{Results on random kernel size}

We expect that the reconstruction quality decays with larger kernel size, as a large kernel size can not consider the local visual feature of the input. 
In the extreme case when the kernel size equals the image dimension, the convolution operator actually combines all pixel values of the input to an output pixel using random weights.
To verify this assumption, we use the rrVGG Conv1\_1 DeConv1\_1 architecture, which simply contains two convolutional operators. 
The random weights are from the $N(0, 0.1)$ distribution. For each kernel size, we randomly generate 30 networks for the reconstruction on 50 sample images as selected above.

\begin{figure}[hbt!]
	\vskip -0.1in
	\centerline{\includegraphics[width=2in]{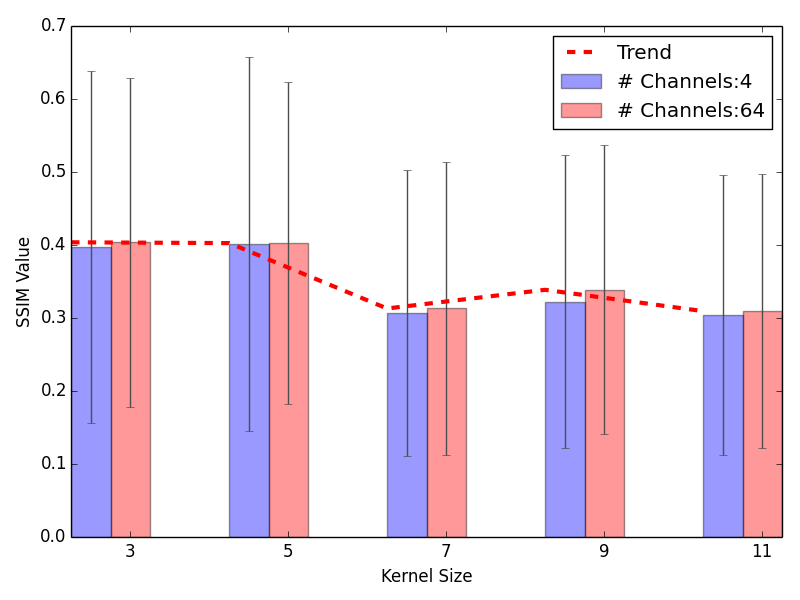}}
	\vskip -0.2in
	\caption{SSIM for various kernel sizes on rrVGG Conv1\_1.}
	\label{FilterSize}
	\vskip -0.2in
\end{figure}

\section{Theories on Random Convolution}
\label{sec:Theory}
In this section, we provide theoretical analysis to explain the surprising empirical results shown in the previous experiments. We will show that for a slight variant of the random CNN architecture, when the number of channels in each layer goes to infinity, the output image will converge to an image highly correlated to the input image. Note that DCN is also a kind of CNN with up-sampling layers, so our result can be directly applied to the CNN-DCN architecture. We first introduce some notations and describe our variant random CNN architecture.

\noindent \textbf{Notations:}
We use \(A_{:,j}\) to denote the $j^{th}$ column vector of matrix $A$ and $\|\mathbf{x}\|$ to denote the $l_2$-norm of vector $\mathbf{x}$. Let \(L\) be the number of layers in the neural network and \(X^{(i)}\in\mathbb{R}^{N_i\times d_i}\) be the feature maps in the \(i^{th}\) layer, where \(N_i\) is the number of channels and \(d_i\) is the dimension of a single channel feature map (i.e. the width of the map times its height). \(X=X^{(0)}\) is the input image and \(\mathbf{f}=X^{(L)}\) is the output image.
\(w^{(i, j)}\), a row vector, is the \(j^{th}\) convolutional filter of the \(i^{th}\) layer if it is a convolutional layer.
We use \(ReLU(\mathbf{x})=\max(\mathbf{x}, 0)\) as the activation function in the following analysis.   

\begin{defn}\label{Def1.1}
	{\textbf{Random CNN architecture} \(\;\)This structure is different from the classic CNN structure in the following three points:
		\vspace{-0.5em}		
		\begin{itemize}
			\item Different filters in the same layer are i.i.d. random vectors and filters in different layers are independent. The probability density function of each filter is isotropic. Let $k_m^{(i)}=\frac{1}{2}\mathbb{E} |w^{(i,j)}_1|^m$ and $K^{(i)}_m=\frac{k^{(i)}_{2m}-(k^{(i)}_m)^2}{(k^{(i)}_m)^2}$.  Suppose that $k_1^{(i)},k_2^{(i)},k_4^{(i)}$ all exist. 
			\item The last layer is the arithmetic mean of the channels of the previous layer, not the weighted combination.
			\item Except for $X^{(L-1)}$, each layer of convolutional feature maps are normalized by a factor of \(\frac{1}{\sqrt{N_i}}\), where \(N_i\) is the number of channels of this layer.
			
		\end{itemize}
	}
\end{defn}

\subsection{Convergence}
As from the previous experiments, we see that when the number of channels increases, the quality of the output image becomes better, so we are interested in what the output image looks like when the number of channels in each convolution layer goes to infinity. We prove that when the number of channels goes to infinity, the output will actually converge. Each pixel in the final convergence value is a constant times the weighted norm of its receptive field. Formally, we state our main results for the convergence value of the random CNN as follows:

\begin{thm}\label{thm1.1}
	{\textbf{(Convergence Value)} \(\;\)Suppose all the pooling layers use $l_2$-norm pooling. When the number of filters in each layer of a random CNN goes to infinity, the output $\mathbf{f}$ corresponding to a fixed input will converge to a fixed image $\mathbf{f}^*$ with probability $1$, where $\mathbf{f}^*=k\mathbf{z}^*$ and $k$ is a constant only related to the CNN architecture and the distribution of random filters and $\mathbf{z}^*_i=\sqrt{\sum_{l \in \mathcal{R}_i} n_{(l,i)} \|X_{:,l}\|^2  }$, where \(\mathcal{R}_i\) is the index set of the receptive field of \(\mathbf{z}^*_i\) and $n_{(l,i)}$ is the number of routes from the $l^{th}$ pixel of a single channel of the input image to the $i^{th}$ output pixel.}
\end{thm}

The proof of \hyperref[thm1.1]{Theorem 5.1} is in the Appendix. Here for the pooling layer, instead of average pooling, which calculates the arithmetic mean, we use $l_2$-norm pooling which calculates the norm of the values in a patch. Intuitively, if most pixels of the input image are similar to their adjacent pixels, the above two pooling methods should have similar outputs. We also show the result for average pooling in the Appendix.

\subsection{Variance}
Now we consider the case of finite number of channels in each layer. We mainly focus on the difference between the real output and the convergence value as a function of the number of channels. We prove that for our random CNN architecture, as the number of channels increases, with high probability, the angle between the real output and the convergence value becomes smaller, which is in accordance with the variant rrVGG experiment results shown in Section 4.2. To give an insight, we first state our result for a two-layer random CNN and then extend it to a multi-layer one.
\begin{thm}\label{thm1.2}
	{\textbf{(Variance)} \(\;\)For a two-layer random CNN with $N$ filters in the first convolutional layer, let $\Theta$ denote the angle between the output $\mathbf{f}$ and the convergence value $\mathbf{f}^*$, then with probability $1-\delta$, $\sin\Theta \le \sqrt{K_1^{(0)} \frac{1}{N \delta}}$}.
\end{thm}

Now we extend \hyperref[thm1.2]{Theorem 5.2} to the multi-layer random CNN setting:
\begin{thm}\label{thm1.3}
	{\textbf{(Multilayer Variance)} \(\;\)Suppose all the pooling layers use $l_2$-norm pooling. For a random CNN with $L$ layers and $N_i$ filters in the $i^{th}$ layer, let $\Theta$ denote the angle between the output $\mathbf{f}$ and the convergence value $\mathbf{f}^*$, suppose that there is at most one route from an arbitrary input pixel to an arbitrary output pixel for simplicity, then with probability $1-\delta$, \[\sin \Theta\le \sqrt{ \frac{L-1}{\overline{N}\delta}}+ \sqrt{ (L-2)\sqrt{\frac{L-1}{\overline{N}\delta}}\prod_{i=0}^{L-2}\lambda_i  }  ,\]
		where $\lambda_i= \frac{1}{\sqrt{1-\frac{\|\mathbf{\epsilon}^{(i)}((\mathbf{z}^{*(i)})^2)\|^2}{\|(\mathbf{z}^{*(i)})^2\|^2 }}}$ and $\frac{1}{\overline{N}}=\frac{1}{L-1}(\frac{K^{(L-2)}_1}{N_{L-1}}+ \sum_{i=1}^{L-2} \frac{K^{(i-1)}_2}{N_i} )$. }
\end{thm}
Here, \(\mathbf{\epsilon}^{(i)}(\mathbf{x})\) actually measures the local similarity of \(\mathbf{x}\). The full definition of \(\mathbf{\epsilon}^{(i)}(\mathbf{x})\) and the proof of \hyperref[thm1.3]{Theorem 5.3} is in the Appendix.

\subsection{Difference on the convergence value and the input}
Finally, we focus on how well our random CNN architecture can reconstruct the original input image. From \hyperref[thm1.2]{Theorem 5.2} and \hyperref[thm1.3]{Theorem 5.3}, we know that with high probability, the angle between the output of a random CNN with finite channels and the convergence value will be upper-bounded. Therefore, to evaluate the performance of reconstruction, we focus on the difference between the convergence value and the input image. We will show that if the input is an image whose pixels are similar to their adjacent pixels, then the angle between the input image \(X\) and the convergence value of the output image will be small. To show the essence more clearly, we state our result for a two-layer random CNN and leave the multi-layer one in the Appendix, which needs more complicated technics but has the same insight as the two-layer one.
\begin{thm}\label{thm1.4}
	{For a two-layer random CNN, suppose each layer has a zero-padding scheme to keep the output dimension equal to the dimension of the original input. The kernel size is \(r\) and stride is \(1\). The input image is \(X\in\mathbb{R}^{d_0}\), which has only one channel, whose entries are all positive. \(\epsilon_{t}=X_{t}-\overline{X_{t}}\) means the difference between one pixel \(X_{t}\) and the mean of the \(r\)-sized image patch whose center is \(X_{t}\). Let $\Phi$ be the angle between the input image \(X\) and the convergence value of the output image, we have  \(\cos\Phi \ge 1-\frac{1}{M}\sum_{t}\epsilon_{t}X_{t}\), where \(M=\sum_{t}X_{t}^2\).}
\end{thm}

The full proof of \hyperref[thm1.4]{Theorem 5.4} is in the Appendix. Note that when the kernel size \(r\) increases, \(\epsilon_t\) will become larger as an image only has local similarity, so that the lower bound of the cosine value becomes worse, which explains the empirical results in Section 4.3.

\section{Discussion and Conclusions}
\label{sec:Conclusion}

In recent years randomization has been attracting increasing attention due to its competitive performance for optimization as compared to fully trained costly models, and the potential for fast deep learning architecture selection. 
In this work, we focus on understanding the inner representation of deep convolutional networks via randomization and deconvolution. 
To our knowledge, this is the first attempt to explore the randomness of convolutional networks by inverting the representation back to the image space through deconvolution and also the first attempt to explore the randomness of the overall CNN-DCN network structure in the literature. 

We first study the representation of random convolutional networks by training a deconvolutional network, and shed some lights on the inner mechanism of CNN. Extensive empirical study shows that the representation of random CNN performs better than that of the pre-trained CNN when used for reconstruction. Our results indicate that the random CNN can retain the photographically accurate information, while the pre-trained CNN discards some irrelevant information for classification, making it harder for the image reconstruction.

When we use random feature extractor for the overall CNN-DCN architecture, with a plenty number of channels in each layer, we can approximately reconstruct the input images. 
We further explore the reconstruction ability of the purely random CNN-DCN architecture from three perspectives: the depth and width of the network, the number of channels and the size of the filter. As compared to the network depth, the dimensions (width) of the convolutional layers contribute significantly to the reconstruction quality. We investigate various number of channels for reconstruction and find that the increasing number of random channels actually promotes the reconstruction quality. For the third point, we verify our assumption that the reconstruction quality will decay with the increasement of the filter size.  

In the end, we provide theoretical foundations to interpret the empirical results on the random CNN-DCN architecture. We prove that for a variant of the random CNN architecture that calculates the average over the feature maps next to the last layer, when the number of channels in each layer goes to infinity, the output image will converge to an image highly correlated to the input. We also bound the error between the convergence value and the actual output when the number of channels in each layer is finite.

\section*{Acknowledgments}
This work was supported by National Natural Science Foundation of China (61602196, 61472147), US Army Research Office (W911NF-14-1-0477).

\appendix
\section{Proof of Theories on Random Convolution}

In the appendix, we prove that for a slight variant of the random CNN architecture, when the number of filters in each layer goes to infinity, the output will converge to a fixed image \(\mathbf{f}^*\) with probability 1 and it is proportional to the square root of the weighted sum of the square of the pixels in the corresponding receptive field; 
when the numbers of filters in each layer are finite, then with high probability, the angle between its output image and \(\mathbf{f}^*\) is bounded.
For random convolutional neural networks with zero-padding in each layer, we also give a lower bound for the cosine value of the angle between the input image and the convergence value of the output image.

For completeness, we repeat the notations and the definition of random CNN architecture.

\noindent \textbf{Notations:}
We use \(A_{:,j}\) to denote the $j^{th}$ column vector of matrix $A$ and use $A_{ij}$ to denote its entry. Let $\mathbf{x}_i$ be the $i^{th}$ entry of vector $\mathbf{x}$. Let \(L\) be the number of layers in the neural network and \(X^{(i)}\in\mathbb{R}^{N_i\times d_i}\) be the feature maps in the \(i^{th}\) layer, where \(N_i\) is the number of channels and \(d_i\) is the dimension of a single channel feature map (i.e. the width of the map times its height). \(X=X^{(0)}\) is the input image and \(\mathbf{f}=X^{(L)}\) is the output image. For convenience, we also define \emph{convolutional feature maps} to be the feature maps after convolutional operation and define \emph{pooled feature maps} and \emph{up-sampled feature maps} in the same way. In the \(i^{th}\) layer, let \(r_i\) be the fixed kernel size or the pool size (e.g. $3\times3$). If \(X^{(i+1)}\) is convolutional feature maps, let \(Y^{(i)}\in\mathbb{R}^{N_ir_i\times \tilde{d_i}}\) be the patched feature maps of $X^{(i)}$, where $\tilde{d_i}$ is the number of patches and in fact $\tilde{d_i}=d_{i+1}$ and \(Y^{(i)}_{:,j}\) is the receptive field of the \(j^{th}\) pixel in a single channel output image after the \(i^{th}\) layer. To form \(Y^{(i)}\), we first divide \(X^{(i)}\) into patches and the $m^{th}$ patch is \(X^{(i)}_{:,\mathcal{D}_m^{(i)} }  \triangleq \{X^{(i)}_{:,j} \;|\;j \in \mathcal{D}_m^{(i)} \}\), where \(\mathcal{D}_m^{(i)}\) is an ordered set of indexes. \(\mathcal{D}_{m,s}^{(i)}\) means the \(s^{th}\) corresponding element in \(\mathcal{D}_{m}^{(i)}, s\in[r_i].\) We assume $\bigcup_{m=1}^{\tilde{d_i}} \mathcal{D}_m^{(i)}=[d_i] $, which is satisfied by most widely used CNNs. By transforming \(X^{(i)}_{:,\mathcal{D}_m^{(i)}}\) into a column vector, we can obtain \(Y^{(i)}_{:,m}\). \(\mathbf{z}^{(i)}\in\mathbb{R}^{\tilde{d_i}}\) is a row vector defined by \(\mathbf{z}^{(i)}_j=\|Y^{(i)}_{:,j}\|\), where $\|\cdot\|$ is the $l_2$-norm operation.  \(w^{(i, j)}\), a row vector, is the \(j^{th}\) convolutional filter of the \(i^{th}\) layer.
If \(X^{(i+1)}\) is pooled feature maps, we also divide \(X^{(i)}\) into patches and \(\mathcal{D}_m^{(i)}\) is the indexes of the \(m^{th}\) patch. If \(X^{(i+1)}\) is up-sampled feature maps we have \(\mathcal{D}_m^{(i)}=\{m\}\), which has only one element. So we can also define $Y^{(i)}$ and $\mathbf{z}^{(i)}$ for pooling and up-sampling layers. For the $j^{th}$ pixel of the output image in the last layer, define its receptive filed on the input image in the first layer as $X_{:,\mathcal{R}_j}=\{X_{:,m}\;|\;m \in \mathcal{R}_j\}$, where $\mathcal{R}_j$ is a set of indexes. The activation function \(ReLU(\mathbf{x})=\max(\mathbf{x}, 0)\) is the element-wise maximum operation between \(\mathbf{x}\) and \(0\) and $(\cdot)^m$ is the element-wise power operation.

\begin{defn}\label{Def1.1a}
	{\textbf{Random CNN architecture.} \(\;\)This structure is different from the classic CNN in the following three points:
		\begin{itemize}
			\item Different filters in the same layer are i.i.d. random vectors and filters in different layers are independent. The probability density function of each filter is isotropic. Let $k_m^{(i)}=\frac{1}{2}\mathbb{E} |w^{(i,j)}_1|^m$ and $K^{(i)}_m=\frac{k^{(i)}_{2m}-(k^{(i)}_m)^2}{(k^{(i)}_m)^2}$.  Suppose $k_1^{(i)},k_2^{(i)},k_4^{(i)}$ all exist. 
			\item The last layer is the arithmetic mean of the channels of the previous layer, not the weighted combination.
			\item Except for $X^{(L-1)}$, each layer of convolutional feature maps are normalized by a factor of \(\frac{1}{\sqrt{N_i}}\), where \(N_i\) is the number of channels of this layer.
			
		\end{itemize}
	}
\end{defn}

\subsection{Convergence}

\begin{thm}\label{Thm 1.1}
	{\textbf{(Convergence Value)} \(\;\)Suppose all the pooling layers use $l_2$-norm pooling. When the number of filters in each layer of a random CNN goes to infinity, the output $\mathbf{f}$ corresponding to a fixed input will converge to a fixed image $\mathbf{f}^*$ with probability $1$, where $\mathbf{f}^*=k\mathbf{z}^*$ and $k$ is a constant only related to the CNN architecture and the distribution of random filters and $\mathbf{z}^*_i=\sqrt{\sum_{l \in \mathcal{R}_i} n_{(l,i)} \|X_{:,l}\|^2  }$, where $n_{(l,i)}$ is the number of routes from the $l^{th}$ input pixel to the $i^{th}$ output pixel.}
\end{thm}
Here for the pooling layer, instead of average pooling, which calculates the arithmatic mean, we use $l_2$-norm pooling which calculates the norm of the values in a patch. We also show the result for average pooling in \hyperref[thm1.7]{Theorem A.7}.

To prove the theorem, we first prove the following lemma.

\begin{lem}\label{Lem 1.2}
	{Suppose $w \in \mathbb{R}^ n, n \ge 2$ is a random row vector and its probability density function is isotropic. $Y \in \mathbb{R}^{n \times d}$ is a constant matrix whose $i^{th}$ column vector is denoted by $y_i$. $\mathbf{z} \in \mathbb{R}^ d$ is a row vector and $\mathbf{z}_i=\|y_i\|$. Let $\mathbf{g} = max \{ wY , 0 \}$. If $k_m = \frac{1}{2}\mathbb{E} |w_1|^m$ exists, then $\mathbb{E} \mathbf{g} ^m = k_m \mathbf{z}^m $} and $\mathbb{E} \mathbf{g}_i \mathbf{g}_j=\frac{1}{\pi}k_2[(\pi-\theta_{ij})\cos\theta_{ij}+\sin\theta_{ij}]\mathbf{z}_i\mathbf{z}_j$, where $\theta_{ij}$ is the angle between $y_i$ and $y_j$.
\end{lem}

\proof{
	Note that $max\{\cdot,\cdot\}$ and $(\cdot)^m$ are both element wise operations. The $i^{th}$ element of $\mathbb{E} \mathbf{g} ^m$ is
	\[(\mathbb{E} \mathbf{g} ^m)_i= \mathbb{E}  max \{ wy_i , 0 \}^m.\]
	Since the probability density function of $w$ is isotropic, we can rotate $y_i$ to $y_i'$ without affecting the value of $\mathbb{E}  max \{ wy_i , 0 \}^m$. Let $y_i'=(\|y_i\|,0,...,0)^T$ , we have
	\begin{align*}
	(\mathbb{E} \mathbf{g} ^m)_i &= \mathbb{E}  max \{ \|y_i\|w_1 , 0 \}^m \\
	&= \mathbf{z}_i^m \mathbb{E}  max \{ w_1 , 0 \}^m \\
	&=\mathbf{z}_i^m \frac{1}{2}\mathbb{E} |w_1^m| \\
	&=k_m \mathbf{z}_i^m.
	\end{align*}
	Where the third equality uses the fact that the marginal distribution of $w_1$ is also isotropic.
	Similarly, we have:
	\[\mathbb{E} \mathbf{g}_i \mathbf{g}_j=\mathbb{E}  max \{ wy_i , 0 \} max \{ wy_j , 0 \}.\]
	We can also rotate $y_i$ and $y_j$ to $y_i'$ and $y_j'$. Let $y_i'=(\|y_i\|,0,0,...,0)^T$ and $y_j'=(\|y_j\|\cos\theta_{ij},\|y_j\|\sin\theta_{ij},0,...,0)^T$ and suppose the marginal probability density function of $(w_1,w_2)$ is $p(\rho)$ which does not depend on $\phi$ since it is isotropic, where $\rho=\sqrt{w_1^2+w_2^2}$ is the radial coordinate and $\phi$ is the angular coordinate. We have:
	\[\mathbb{E} \mathbf{g}_i \mathbf{g}_j=\mathbf{z}_i\mathbf{z}_j\int_{0}^{\infty} p(\rho) \rho^3 d \rho \int_{\theta_{ij}-\frac{\pi}{2}}^{\frac{\pi}{2}}\cos(\theta_{ij}-\phi)\cos\phi d \phi.\]
	Note that:
	\[\int_{0}^{\infty} p(\rho) \rho^3 d \rho = \frac{1}{2\pi}\mathbb{E} \rho^2=\frac{1}{\pi}\mathbb{E} w_1^2 =\frac{2}{\pi}k_2,\]
	\[\int_{\theta_{ij}-\frac{\pi}{2}}^{\frac{\pi}{2}}\cos(\theta_{ij}-\phi)\cos\phi d \phi=\frac{1}{2}((\pi-\theta_{ij})\cos\theta_{ij}+\sin\theta_{ij}).\]
	We obtain the second part of this lemma.
	\qed}

Now, we come to proof of \hyperref[Thm 1.1]{Theorem A.1}.
\proof{According to \hyperref[Lem 1.2]{Lemma A.2}, if $X^{(i+1)}$ is convolutional feature maps, we can directly obtain:
	\begin{align*}		
	\mathbb{E} (X^{(i+1)}_{1,:})^2 
	&=\frac{1}{N_{i+1}}\mathbb{E} max\{w^{(i,1)}Y^{(i)},0\}^2\\
	&=\frac{k_2^{(i)}}{N_{i+1}} (\mathbf{z}^{(i)})^2, \;\;\; 0 \le i \le L-2 
	\end{align*}				
	where we have fixed $ Y^{(i)}$ and the expectation is taken over random filters in the $i^{th}$ layer only. Since different channels in $X^{(i+1)}$ are i.i.d. random variables, according to the strong law of large numbers, we have:
	\[\sum_{j=1}^{N_{i+1}} (X^{(i+1)}_{j,:})^2 \stackrel{a.s.}{\longrightarrow} k_2^{(i)}(\mathbf{z}^{(i)})^2 \;\;\;\;\;\;\; \text{when} \; N_{i+1} \rightarrow \infty,  \]
	which implies that with probability $1$,
	\begin{align*}			
	\lim_{N_{i+1} \rightarrow \infty} (\mathbf{z}^{(i+1)}_m)^2 
	&=\lim_{N_{i+1} \rightarrow \infty} \sum_{l \in \mathcal{D}_m^{(i+1)}} \sum_{j=1}^{N_{i+1}} (X^{(i+1)}_{j,l})^2 \\
	&= k_2^{(i)} \sum_{l \in \mathcal{D}_m^{(i+1)}}(\mathbf{z}^{(i)}_l)^2.  
	\end{align*}		
	Suppose that all $N_j$ for $1 \le j \le i$ have gone to infinity and $\mathbf{z}^{(i)}$ has converged to $\mathbf{z}^{*(i)}$, the above expression is the recurrence relation between $\mathbf{z}^{*(i+1)}$ and $\mathbf{z}^{*(i)}$ in a convolutional layer: $(\mathbf{z}^{*(i+1)}_m)^2  = k_2^{(i)} \sum_{l \in \mathcal{D}_m^{(i+1)}}(\mathbf{z}^{*(i)}_l)^2.  $
	
	If $X^{(i+1)}$ is $l_2$-norm pooled feature maps, we have $\|X^{(i+1)}_{:,l}\|=\mathbf{z}^{(i)}_l$ by defination. Therefore, $$(\mathbf{z}^{(i+1)}_m)^2=\sum_{l \in \mathcal{D}_m^{(i+1)}}(\mathbf{z}^{(i)}_l)^2.$$
	
	If $X^{(i+1)}$ is up-sampled feature maps, a pixel $X^{(i)}_{jp}$ will be up-sampled to a $r$-sized block $\{X^{(i+1)}_{jp_q}\;|\;q \in [r]\}$, where $X^{(i+1)}_{jp_1}=X^{(i)}_{jp}$ and all the other elements are zeros. Define $\tilde{\mathcal{D}}^{(i+1)}_m=\{p\;|\;p_1 \in \mathcal{D}^{(i+1)}_m\}$, we have:
	$$(\mathbf{z}^{(i+1)}_m)^2=\sum_{l \in \tilde{\mathcal{D}}_m^{(i+1)}}(\mathbf{z}^{(i)}_l)^2.$$
	
	So far, we have obtained the recurrence relation in each layer. In order to get $\mathbf{z}^{*(i+1)}$ given $\mathbf{z}^{*(i)}$, we use the same sliding window scheme on $\mathbf{z}^{*(i)}$ as that of the convolutional, pooling or upsampling operation on the feature maps. The only difference is that in a convolutional layer, instead of calculating the inner product of a filter and the vector in a sliding window, we simply calculate the $l_2$-norm of the vector in the sliding window and then multiply it by $\sqrt{k_2^{(i)}}$. Note that $\mathbf{z}^{*(0)}$ can be directly obtained from the input image. Repeat this process layer by layer and we can obtain $\mathbf{z}^{*(L-2)}$.
	
	\noindent According to \hyperref[Lem 1.2]{Lemma A.2}, we have:
	\[\mathbb{E} X^{(L-1)}_{1,:}=\mathbb{E} \max \{w^{(L-2,1)}Y^{(L-2)},0\} =k_1^{(L-2)} \mathbf{z}^{(L-2)}. \]
	Suppose that $\mathbf{z}^{(L-2)}$ has converged to $\mathbf{z}^{*(L-2)}$, and by \hyperref[Def1.1a]{Definition A.1}, $\mathbf{f} = \frac{1}{N_{L-1}}\sum_{i=1}^{N_{L-1}} X^{(L-1)}_{i,:}$, we have:
	\[\mathbf{f} \stackrel{a.s.}{\longrightarrow}k_1^{(L-2)} \mathbf{z}^{*(L-2)} \;\;\;\;\;\;\; \text{when} \;N_i \rightarrow \infty,\;\;\;i \in [L-1]. \]
	Let $k=k_1^{(L-2)}  \prod_{i=0}^{L-3} k_{2}^{(i)}$ and $\mathbf{z}^{*}=(\prod_{i=0}^{L-3} k_{2}^{(i)})^{-1}\mathbf{z}^{*(L-2)}$, we have $\mathbf{f}^*=k\mathbf{z}^*$. Note that $\mathbf{z}^{*}$ is obtained through a multi-layer sliding window scheme similar to the CNN structure. It only depends on the input image and the scheme. It is easy to verify that $\mathbf{z}^*_i$ is the square root of the weighted sum of the square of input pixel values within the receptive field of the $i^{th}$ output pixel, where the weight of an input image pixel is the number of routes from it to the output pixel. \qed}

\subsection{Variance}

\begin{thm}\label{Thm 1.3}
	{\textbf{(Variance)} \(\;\)For a two-layer random CNN with $N$ filters in the first convolutional layer, let $\Theta$ denote the angle between the output $\mathbf{f}$ and the convergence value $\mathbf{f}^*$, then with probability $1-\delta$, $\sin\Theta \le \sqrt{K_1^{(0)} \frac{1}{N \delta}}$}.
\end{thm}

\proof{According to \hyperref[Thm 1.1]{Theorem A.1}, we have $\mathbb{E}\mathbf{f}=\mathbf{f}^*=k_1^{(0)} \mathbf{z}^{*}$, where $\mathbf{z}^{*}=\mathbf{z}^{*(0)}$. For a two-layer CNN, we can directly obtain:
	\[\mathbf{f}=\frac{1}{N} \sum_{i=1}^{N} X^{(1)}_{i,:}=\frac{1}{N} \sum_{i=1}^{N} max\{w^{(0,i)} Y^{(0)},0\},\]
	Since different channels are i.i.d. random variables, we have $\mathbb{E}X^{(1)}_{i,:}=\mathbb{E}\mathbf{f}=k_1^{(0)} \mathbf{z}^{*}$. Then,
	\begin{align*}
	\mathbb{E} \|\mathbf{f}-\mathbb{E}\mathbf{f}\|^2&=
	\frac{1}{N}\mathbb{E} \|X^{(1)}_{i,:}-\mathbb{E}X^{(1)}_{i,:}\|^2\\ &=\frac{1}{N}(\mathbb{E}\|X^{(1)}_{i,:}\|^2-\|\mathbb{E}X^{(1)}_{i,:}\|^2) \\
	&=\frac{1}{N} (\sum_{j=1}^{\tilde{d_0}} \mathbb{E} \max\{w^{(0,i)} Y^{(0)}_{:,j},0\}^2-(k_1^{(0)})^2 \|\mathbf{z}^*\|^2)	\\
	&=\frac{1}{N} (\sum_{j=1}^{\tilde{d_0}} k_2^{(0)} \|Y^{(0)}_{:,j}\|^2-(k_1^{(0)})^2 \|\mathbf{z}^*\|^2)	\\	
	&=\frac{1}{N} (k_2^{(0)}-(k_1^{(0)})^2) \|\mathbf{z}^*\|^2
	\end{align*}
	According to Markov inequality, we have:
	\[Pr(\|\mathbf{f}-\mathbb{E}\mathbf{f}\|^2 \ge \epsilon^2) \le \frac{1}{N\epsilon^2} (k_2^{(0)}-(k_1^{(0)})^2) \|\mathbf{z}^*\|^2.\]
	Let $\delta=\frac{1}{N\epsilon^2} (k_2^{(0)}-(k_1^{(0)})^2) \|\mathbf{z}^*\|^2$, then with probability $1-\delta$:
	\begin{align*}		
	\sin\Theta &\le \frac{\|\mathbf{f}-\mathbb{E}\mathbf{f}\|}{\|\mathbb{E}\mathbf{f}\|} \le \frac{\epsilon}{k_1^{(0)}\|\mathbf{z}^*\|}\\
	&= \sqrt{\frac{k_2^{(0)}-(k_1^{(0)})^2}{(k_1^{(0)})^2} \frac{1}{N \delta}}=\sqrt{K_1^{(0)} \frac{1}{N \delta}}.    
	\end{align*}		
	\qed
}

To extend the above two-layer result to a multi-layer one, we first prove the following lemma. Note that in this lemma, $\mathcal{D}^{(i)}$ should be replaced by $\tilde{\mathcal{D}}^{(i)}$ defined in the proof of \hyperref[Thm 1.1]{Theorem A.1} if $X^{(i)}$ is up-sampled feature maps.

\begin{lem}\label{Lem 1.4}
	{Let $\phi^{(i)}(\cdot):\mathbb{R}^{\tilde{d}_{i}} \rightarrow \mathbb{R}^{\tilde{d}_{i+1}} $ for $0 \le i \le L-3$ denote the recurrence relation of $\{(\mathbf{z}^{*(i)})^2\}$ (i.e. $(\mathbf{z}^{*(i+1)})^2=\phi^{(i)}((\mathbf{z}^{*(i)})^2)$), then $\phi^{(i)} (\cdot)$ is a linear mapping. For simplicity, suppose that for any $i \in [L-3]$, $card(\mathcal{D}_m^{(i)})=r_i$ for any $m \in [\tilde{d}_i]$. And $ \mathcal{D}_l^{(i)} \bigcap \mathcal{D}_m^{(i)}=\varnothing$ for any $l,m \in [\tilde{d}_i]$ if $l \ne m$. Then we can obtain $\|\phi^{(i)}(\mathbf{x})\|^2 = r_{i+1}(k^{(i)})^2(\|\mathbf{x}\|^2-\|\mathbf{\epsilon}^{(i)}(\mathbf{x})|^2)  \le r_{i+1}(k^{(i)})^2\|\mathbf{x}\|^2$, where $k^{(i)}=k_2^{(i)}$ for convolutional layers and $k^{(i)}=1$ for pooling and up-sampling layers and $\mathbf{\epsilon}^{(i)}(\cdot)$ is defined by $\mathbf{\epsilon}^{(i)}(\mathbf{x})_j=\mathbf{x}_j-\frac{1}{r_{i+1}}\sum_{l \in \mathcal{D}_m^{(i+1)}}\mathbf{x}_l$, where $m$ satisfies $j \in \mathcal{D}_m^{(i+1)}$ .  }
\end{lem}

\proof{According to the definition of $\phi^{(i)}(\cdot)$ and \hyperref[Thm 1.1]{Theorem A.1}, we have:
	\[\phi^{(i)}(\mathbf{x})_m=k^{(i)}\sum_{j \in \mathcal{D}_m^{(i+1)}}\mathbf{x}_j.\]
	
	It is easy to verify that for any $c \in \mathbb{R}$ and $\mathbf{x},\mathbf{y} \in \mathbb{R}^{d_{i+1}}$ we have $\phi^{(i)}(c\mathbf{x})=c\phi^{(i)}(\mathbf{x})$ and $\phi^{(i)}(\mathbf{x}+\mathbf{y})=\phi^{(i)}(\mathbf{x})+\phi^{(i)}(\mathbf{y})$. So $\phi^{(i)} (\cdot)$ is a linear mapping.
	
	\noindent Define $\overline{\mathbf{x}}_m=\frac{1}{r_{i+1}}\sum_{j \in \mathcal{D}_m^{(i+1)}}\mathbf{x}_j $, which is the average value of the $m^{th}$ patch. Let $\mathbf{\epsilon}^{(i)}(\mathbf{x})_j=\mathbf{x}_j-\overline{\mathbf{x}}_m$, where $m$ satisfies $j \in \mathcal{D}_m^{(i+1)}$. We have:
	\begin{align*}
	\sum_{j \in \mathcal{D}_m^{(i+1)}}\mathbf{x}_j^2 &= \sum_{j \in \mathcal{D}_m^{(i+1)}}(\overline{\mathbf{x}}_m+\mathbf{\epsilon}^{(i)}(\mathbf{x})_j)^2\\
	&=r_{i+1}\overline{\mathbf{x}}_m^2+ \sum_{j \in \mathcal{D}_m^{(i+1)}}\mathbf{\epsilon}^{(i)}(\mathbf{x})_j^2,
	\end{align*}
	Since $\{\mathcal{D}_m^{(i+1)} | m \in [\tilde{d}_{i+1}] \}$ is a partition of $[\tilde{d}_{i}]$ under our assumptions, we have $\|\mathbf{x}\|^2=\sum_{m=1}^{\tilde{d}_{i+1}} \sum_{j \in \mathcal{D}_m^{(i+1)}}\mathbf{x}_j^2 $ and $ \|\phi^{(i)}(\mathbf{x})\|^2=r_{i+1}^2 (k^{(i)})^2 \sum_{m=1}^{\tilde{d}_{i+1}} \overline{\mathbf{x}}_m^2$, which implies that
	\begin{align*}	
	\|\phi^{(i)}(\mathbf{x})\|^2 &= r_{i+1}(k^{(i)})^2(\|\mathbf{x}\|^2-\|\mathbf{\epsilon}^{(i)}(\mathbf{x})|^2)  \\
	&\le r_{i+1}(k^{(i)})^2\|\mathbf{x}\|^2.  
	\end{align*}	
	\qed
}
\begin{thm}\label{Thm 1.5}
	{\textbf{(Multilayer Variance)} \(\;\)Suppose all the pooling layers use $l_2$-norm pooling. For a random CNN with $L$ layers and $N_i$ filters in the $i^{th}$ layer, let $\Theta$ denote the angle between the output $\mathbf{f}$ and the convergence value $\mathbf{f}^*$, suppose that there is at most one route from an arbitrary input pixel to an arbitrary output pixel for simplicity, then with probability $1-\delta$, \[\sin \Theta\le \sqrt{ \frac{L-1}{\overline{N}\delta}}+ \sqrt{ (L-2)\sqrt{\frac{L-1}{\overline{N}\delta}}\prod_{i=0}^{L-2}\lambda_i  } ,\]
		where $\lambda_i= \frac{1}{\sqrt{1-\frac{\|\mathbf{\epsilon}^{(i)}((\mathbf{z}^{*(i)})^2)\|^2}{\|(\mathbf{z}^{*(i)})^2\|^2 }}}$ and \\
		\centerline{ $\frac{1}{\overline{N}}=\frac{1}{L-1}(\frac{K^{(L-2)}_1}{N_{L-1}}+ \sum_{i=1}^{L-2} \frac{K^{(i-1)}_2}{N_i} )$. }}
\end{thm}
\proof{We will bound $\Theta$ recursively. Suppose that the angle between $(\mathbf{z}^{(i)})^2$ and $(\mathbf{z}^{*(i)})^2$ is $\theta_i$. We have $\theta_0=0$. Let $\mathbf{g}^{(i+1,j)}=(X^{(i+1)}_{j,:})^2 $ and $\mathbf{g}^{(i+1)} = \frac{1}{N_{i+1}}\sum_{j=1}^{N_{i+1}} \mathbf{g}^{(i+1,j)}$. If $X^{(i+1)}$ is convolutional feature maps, we have obtained in the proof of \hyperref[Thm 1.1]{Theorem A.1} that $\mathbb{E} \mathbf{g}^{(i+1)}=\mathbb{E} \mathbf{g}^{(i+1,j)} = k_2^{(i)}(\mathbf{z}^{(i)})^2.$ Using similar method to the proof of \hyperref[Thm 1.3]{Theorem A.3} and let $\alpha_{i+1}$ denote the angle between $\mathbf{g}^{(i+1)}$ and $\mathbb{E} \mathbf{g}^{(i+1)}$, we can derive that with probability $1-\delta_{i+1}$, 
	\[\sin \alpha_{i+1} \le \sqrt{K^{(i)}_2 \frac{1}{N_{i+1} \delta_{i+1}}}. \]
	For a $l_2$-norm pooling layer or an up-sampling layer, we have:
	\[\sin \alpha_{i+1} =0 \le \sqrt{K^{(i)}_2 \frac{1}{N_{i+1} \delta_{i+1}}}. \]
	Let $\beta_{i+1}$ denote the angle between  $\mathbf{g}^{(i+1)}$ and $(\mathbf{z}^{*(i)})^2$. We have:
	\[\sin \beta_{i+1} \le \sin(\theta_{i}+\alpha_{i+1}) \le  \sin\theta_{i}+ \sin\alpha_{i+1}.  \]
	Note that there exists a constant $\gamma$ such that $\sin \beta_{i+1}=\frac{\|\gamma\mathbf{g}^{(i+1)}- (\mathbf{z}^{*(i)})^2\|}{\|(\mathbf{z}^{*(i)})^2\|}$. In fact, we can find the value of $\gamma$ is $\frac{(\mathbf{z}^{*(i)})^2(\mathbf{g}^{(i+1)})^T}{\|\mathbf{g}^{(i+1)}\|^2}$, where $(\cdot)^T$ means transpose. 
	We use $\phi^{(i)} (\cdot)$ to denote the recurrence relation of $\{(\mathbf{z}^{*(i)})^2\}$ (i.e. $(\mathbf{z}^{*(i+1)})^2=\phi^{(i)}((\mathbf{z}^{*(i)})^2)$). Note that $(\mathbf{z}^{(i+1)})^2=\phi^{(i)}(\mathbf{g}^{(i+1)})$ and $\phi^{(i)}(\cdot)$ is linear, using the result in \hyperref[Lem 1.4]{Lemma A.4}, we can obtain:
	\begin{align*}
	\sin \theta_{i+1} &\le \frac{\|\gamma (\mathbf{z}^{(i+1)})^2- (\mathbf{z}^{*(i+1)})^2\|}{\|(\mathbf{z}^{*(i+1)})^2\|}	\\
	&=	\frac{\|\phi^{(i)}(\gamma \mathbf{g}^{(i+1)}- (\mathbf{z}^{*(i)})^2)\|}{\|\phi^{(i)}((\mathbf{z}^{*(i)})^2)\|}\\
	&\le \frac{\|\gamma \mathbf{g}^{(i+1)}- (\mathbf{z}^{*(i)})^2\|}{\sqrt{\|(\mathbf{z}^{*(i)})^2\|^2-\|\mathbf{\epsilon}^{(i)}((\mathbf{z}^{*(i)})^2)\|^2}}\\
	&=\lambda_i\sin \beta_{i+1}\\
	& \le \lambda_i(\sin\theta_{i}+ \sin\alpha_{i+1}).
	\end{align*}
	Where we defined $\lambda_i= \frac{1}{\sqrt{1-\frac{\|\mathbf{\epsilon}^{(i)}((\mathbf{z}^{*(i)})^2)\|^2}{\|(\mathbf{z}^{*(i)})^2\|^2 }}}$ for $0 \le i \le L-3$, which is usually slightly bigger than $1$ if the input is a natural image. 
	
	\noindent So far, we have derived the recurrence relation between $\sin \theta_{i+1}$ and $\sin \theta_{i}$. So we can get the bound of $\theta_{L-2}$.
	However, note that $\theta_{L-2}$ is the angle between $(\mathbf{z}^{(L-2)})^2$ and $(\mathbf{z}^{*(L-2)})^2$ instead of that between $\mathbf{z}^{(L-2)}$ and $\mathbf{z}^{*(L-2)}$. We will denote the latter by $\mu$ which is what we really need.
	\noindent We know that there exists a constant $\gamma'$ such that $\sin \theta_{L-2}=\frac{\|\gamma'(\mathbf{z}^{(L-2)})^2- (\mathbf{z}^{*(L-2)})^2\|}{\|(\mathbf{z}^{*(L-2)})^2\|}$. Note that for any $\mathbf{a},\mathbf{b} \in {\mathbb{R}^+}^n$, according to Cauchy-Schwarz inequality, we have:
	\begin{align*}		
	n \sum_{i=1}^n (\mathbf{a}_i^2-\mathbf{b}_i^2)^2 &\ge (\sum_{i=1}^n |\mathbf{a}_i^2-\mathbf{b}_i^2|)^2 \\
	&= (\sum_{i=1}^n |\mathbf{a}_i-\mathbf{b}_i|(\mathbf{a}_i+\mathbf{b}_i))^2 \\
	&\ge (\sum_{i=1}^n (\mathbf{a}_i-\mathbf{b}_i)^2)^2, 
	\end{align*}		
	which implies that $\|\mathbf{a}-\mathbf{b}\|^4 \le n\|\mathbf{a}^2-\mathbf{b}^2\|^2$. Then we can bound $\mu$:
	\begin{align*}
	\sin \mu &\le \frac{\|\sqrt{\gamma'}(\mathbf{z}^{(L-2)})- (\mathbf{z}^{*(L-2)})\|}{\|\mathbf{z}^{*(L-2)}\|} \\
	&\le \sqrt{\frac{\sqrt{\tilde{d}_{L-2}}\|(\mathbf{z}^{*(L-2)})^2\| }{\|\mathbf{z}^{*(L-2)}\|^2} \sin \theta_{L-2}} 
	\end{align*}
	If we define $\tilde{d}_{L-1}=1$ and $\mathcal{D}_1^{(L-1)}=[d_{L-1}]$, then we can define $\phi^{L-2}(\cdot): \mathbb{R}^{d_{L-1}} \rightarrow \mathbb{R}$ as $\phi^{(L-2)}(\mathbf{x})=\sum_{j \in [d_{L-1}]} \mathbf{x}_j$. Note that $\|\mathbf{z}^{*(L-2)}\|^2=\phi^{L-2}((\mathbf{z}^{*(L-2)})^2)$ by definition. Then according to \hyperref[lem 1.2]{Lemma A.2}, let $\lambda_{L-2}= \frac{1}{\sqrt{1-\frac{\|\mathbf{\epsilon}^{(L-2)}((\mathbf{z}^{*(L-2)})^2)\|^2}{\|(\mathbf{z}^{*(L-2)})^2\|^2 }}}$, we have
	\[\sin\mu \le \sqrt{\lambda_{L-2}\sin\theta_{L-2}}  \]
	Let $v$ denote the angle between  $\mathbf{z}^{(L-2)}$ and $\mathbf{f}$, we have obtained its bound in \hyperref[Thm 1.3]{Theorem A.3}. With probability $1-\delta_{L-1}$, we have $\sin v \le \sqrt{K^{(L-2)}_1 \frac{1}{N_{L-1} \delta_{L-1}}}$. 
	\noindent With all the bounds above, define $\overline{N}$ by $\frac{1}{\overline{N}}=\frac{1}{L-1}(\frac{K^{(L-2)}_1}{N_{L-1}}+ \sum_{i=1}^{L-2} \frac{K^{(i-1)}_2}{N_i} ) $ and choose $\delta_i=\frac{\delta\overline{N}K^{(i)}_2}{(L-1)N_i}$ for $i \le L-2$ and $\delta_{L-1}=
	\frac{\delta\overline{N}K^{(L-2)}_1}{(L-1)N_{L-1}}$ for simplicity, we can obtain the bound of $\Theta$: with probability $1-\delta$,
	\[\sin \Theta\le \sqrt{ \frac{L-1}{\overline{N}\delta}}+ \sqrt{ (L-2)\sqrt{\frac{L-1}{\overline{N}\delta}}\prod_{i=0}^{L-2}\lambda_i  } \] \qed
}

\subsection{Difference of the convergence value and the input}

\noindent For this part, we will give a detailed proof for a two-layer CNN and argue that the result can be directly extended to a multi-layer one only with a few slight changes of definition.

\begin{thm}\label{Thm 1.6}
	{\(\;\)For a two-layer random CNN, suppose that each layer has a zero-padding scheme to keep the output dimension equal to the dimension of the original input. The kernel size is \(r\) and stride is \(1\). The input image is \(X\in\mathbb{R}^{d_0}\), whose entries are all positive. \(\epsilon_{t}=X_{t}-\overline{X_{t}}\) means the difference between one pixel \(X_{t}\) and the mean of the \(r\)-sized image patch whose center is \(X_{t}\). Let $\Phi$ be the angle between the input image \(X\) and the convergence value of the output image, we have  \(\cos\Phi \ge 1-\frac{1}{M}\sum_{t}\epsilon_{t}X_{t}\), where \(M=\sum_{t}X_{t}^2\).}
\end{thm}

\proof{According to \hyperref[Thm 1.1]{Theorem A.1}, we know that \(\mathbf{f}^*=k\mathbf{z}^*\), where \(\mathbf{z}_{t}\) is the square root of the sum of the square of the pixels in its corresponding receptive field (i.e. the $l_2$-norm of the pixels), which means \(\mathbf{z}_{t}=\sqrt{\sum_{\alpha\in\mathcal{R}_t}X_{\alpha}^2}\) and $k$ is a constant related to the network structure and distribution of the random filters. With zero-padding on the input image, we can calculate the angle between \(\mathbf{f}^*\) and \(X\):
	\begin{align*}
	\cos \Phi = \frac{\sum_{t}\left(\sqrt{\sum_{\alpha\in\mathcal{R}_t}X_{\alpha}^2}X_{t}\right)}{\sqrt{\sum_{t}\sum_{\alpha\in\mathcal{R}_t}X_{\alpha}^2}\sqrt{\sum_{t}X_{t}^2}}.
	\end{align*}	
	As each pixel contributes at most \(r\) times to the final output, we have $$\sqrt{\sum_{t}\sum_{\alpha\in\mathcal{R}_t}X_{\alpha}^2}\le \sqrt{r}\sqrt{M}.$$
	Also, using Cauchy-Schwarz Inequality and the fact that all \(X_{t}\) are positive, we can obtain
	$$\sqrt{\sum_{\alpha\in\mathcal{R}_t}X_{\alpha}^2}\ge \frac{1}{\sqrt{r}}\sum_{\alpha\in\mathcal{R}_t}X_{\alpha}.$$
	Now, we can bound the above \(\cos \Phi\) as follows:
	\begin{align*}
	\cos \Phi &\ge \frac{\sum_{t}\left(\frac{1}{\sqrt{r}}X_{t}\sum_{\alpha\in\mathcal{R}_t}X_{\alpha}\right)}{\sqrt{r}\sqrt{M}\sqrt{M}} \\
	&= \frac{1}{M}\sum_{t}X_{t}\overline{X_{t}} \\
	&= \frac{1}{M}\sum_{t}X_{t}(X_{t}-\epsilon_{t}) \\ 
	&= 1-\frac{1}{M}\sum_{t}\epsilon_{t}X_{t}.
	\end{align*} \qed}

\noindent The above theorem indicates that, if the input is an image whose pixels are similar to their adjacent pixels, then the the angle between the input image \(X\) and the convergence value of the output image will be small.\\

\noindent We point out that the above theorem can be directly extended to multi-layer convolutional neural networks. Suppose that the neural network has multiple layers. According to \hyperref[Thm 1.1]{Theorem A.1}, \(\mathbf{f}^*=k\mathbf{z}^*\). Now, the pixels in the receptive fields contribute unequally to the corresponding output. Note that when the network is multi-layer, the receptive field is greatly enlarged. We can similarly obtain:
\begin{align*}
\cos \Phi &= \frac{\sum_{t}\left(\sqrt{\sum_{\alpha\in \mathcal{R}_{t}}n_{(\alpha, t)} X_{\alpha}^2}X_{t}\right)}{\sqrt{\sum_{t}\sum_{\alpha\in \mathcal{R}_{t}}n_{(\alpha, t)} X_{\alpha}^2}\sqrt{\sum_{t}X_{t}^2}}.
\end{align*}

\noindent Here, \(\mathcal{R}_{t}\) is the index set of the receptive field of \(\mathbf{f}_{t}\) and \(n_{(\alpha, t)}\) is the number of routes from \(X_\alpha\) to \(\mathbf{f}_{t}\). Suppose that the receptive field of each \(\mathbf{f}_t\) has the same size and shape, \(X_t\) is at a fixed relative position of the receptive field of \(\mathbf{f}_t\) and \(n_{(\alpha, t)}\) only depends on the relative position between \(X_{\alpha}\) and \(X_t\). Let \(\overline{X_t}=\frac{\sum_{\alpha\in\mathcal{R}_t}n_{(\alpha, t)}X_\alpha}{\sum_{\alpha\in\mathcal{R}_t}n_{(\alpha, t)}}\) be the weighted average and \(\epsilon_t=X_t-\overline{X_t}\). By using the same technique above, we can obtain that
$$\cos \Phi \ge 1-\frac{1}{M}\sum_t\epsilon_tX_t.$$

\noindent Note that although the bound is the same as the two-layer convolutional neural network, as the receptive field is enlarged, \(\epsilon_{t}\) can be much larger, so that the above bound will be worse.

We also give the convergence value for average pooling in the next theorem.
\begin{thm}\label{thm1.7}
	{\textbf{(Convergence Value, average pooling)} \(\;\)Suppose all the pooling layers use average pooling. When the number of filters in each layer of a random CNN goes to infinity, the output $\mathbf{f}$ corresponding to a fixed input will converge to a fixed image $\mathbf{f}^*$ with probability $1$.
	}
\end{thm}

\proof{Define $C^{(i)} \in \mathbb{R}^{d_i \times d_i}$ by  $C^{(i)}_{jk} = (X^{(i)}_{:,j})^T X^{(i)}_{:,k}$. If $X^{(i+1)}$ is convolutional feature maps, according to \hyperref[Lem 1.2]{Lemma A.2}, we have:
	\begin{align*}		
	& \mathbb{E} X^{(i+1)}_{1,j}X^{(i+1)}_{1,k} \\
	&=\frac{1}{N_{i+1}}\mathbb{E} max\{w^{(i,1)}Y^{(i)}_{:,j},0\}max\{w^{(i,1)}Y^{(i)}_{:,k},0\}\\
	&=\frac{1}{N_{i+1}}k_2^{(i)}h(\varphi^{(i)}_{jk})\mathbf{z}^{(i)}_j\mathbf{z}^{(i)}_k,
	\end{align*}		
	where $\varphi^{(i)}_{jk}$ is the angle between $Y^{(i)}_{:,j}$ and $Y^{(i)}_{:,k}$ and we defined $h(x)=\frac{1}{\pi}[(\pi-x)\cos x+\sin x]$ for abbreviation. We have fixed $ Y^{(i)}$ and the expectation is taken over random filters in the $i^{th}$ layer only. Since different channels in $X^{(i+1)}$ are i.i.d. random variables, according to the strong law of large numbers, we have:
	\begin{align*}
	C^{(i+1)}_{jk} &= \sum_{l=1}^{N_{i+1}} X^{(i+1)}_{l,j} X^{(i+1)}_{l,k} \stackrel{a.s.}{\longrightarrow} k_2^{(i)}h(\varphi^{(i)}_{jk})\mathbf{z}^{(i)}_j\mathbf{z}^{(i)}_k \;\;\;\;\;\;\; \\
	& \text{when} \; N_{i+1} \rightarrow \infty. 
	\end{align*}
	Note that:
	\[\mathbf{z}^{(i)}_j= \sqrt {\sum_{l \in \mathcal{D}^{(i)}_j} \|X^{(i) }_{:,l}\|^2}= \sqrt {\sum_{l \in \mathcal{D}^{(i)}_j} C^{(i)}_{ll} },\]
	\[\cos \varphi^{(i)}_{jk}=\frac{(Y^{(i)}_{:,j})^TY^{(i)}_{:,k}}{\mathbf{z}^{(i)}_j\mathbf{z}^{(i)}_k}=\frac{\sum_{s=1}^{r_i} C^{(i)}_{ \mathcal{D}^{(i)}_{j,s} \mathcal{D}^{(i)}_{k,s}} }{\mathbf{z}^{(i)}_j\mathbf{z}^{(i)}_k}. \]
	Suppose that all $N_j$ for $1 \le j \le i$ have gone to infinity and $C^{(i)}$ has converged to $C^{*(i)}$, the above expressions are the recurrence relation between $C^{*(i+1)}$ and $C^{*(i)}$ for a convolutional layer.
	If $X^{(i+1)}$ is average-pooled feature maps, we have:
	\[X^{(i+1)}_{:,j} =\frac{1}{r_i} \sum_{l \in \mathcal{D}_j^{(i)}} X^{(i)}_{:,l}. \]
	We have:
	\begin{align*}		
	C^{(i+1)}_{jk} &=(X^{(i+1)}_{:,j})^TX^{(i+1)}_{:,k}\\
	&=\frac{1}{r_i^2}\sum_{l \in \mathcal{D}_j^{(i)}, m \in \mathcal{D}_k^{(i)}} (X^{(i)}_{:,l})^TX^{(i)}_{:,m}\\
	&=\frac{1}{r_i^2}\sum_{l \in \mathcal{D}_j^{(i)}, m \in \mathcal{D}_k^{(i)}}C^{(i)}_{lm},
	\end{align*}		 
	which is the recurrence relation for an average pooling layer. 
	
	For an up-sampling layer, a pixel $X^{(i)}_{jk}$ will be up-sampled to a block $\{X^{(i+1)}_{jk_m}\; | \;m \in [r] \}$, where $X^{(i+1)}_{jk_1}=X^{(i)}_{jk}$ and all the other elements are zeros. We have:
	\begin{equation*}
	C^{(i+1)}_{j_lk_m}=\begin{cases}
	C^{(i)}_{jk} & \text{for }l=m=1, \\
	0 & \text{otherwise}.
	\end{cases}
	\end{equation*}
	Note that we can directly calculate $C^{*(0)}$ according to the input image. So we can recursively obtain $C^{*(L-2)}$ and thus $\mathbf{z}^{*(L-2)}$.
	
	\noindent According to \hyperref[Lem 1.2]{Lemma A.2}, we have:
	\[\mathbb{E} X^{(L-1)}_{1,:}=\mathbb{E} \max \{w^{(L-2,1)}Y^{(L-2)},0\} =k_1 \mathbf{z}^{(L-2)}. \]
	Suppose that $\mathbf{z}^{(L-2)}$ has converged to $\mathbf{z}^{*(L-2)}$, and by \hyperref[Def1.1a]{Definition A.1}, $\mathbf{f} = \frac{1}{N_{L-1}}\sum_{i=1}^{N_{L-1}} X^{(L-1)}_{i,:}$, we have:
	\[\mathbf{f} \stackrel{a.s.}{\longrightarrow}k_1 \mathbf{z}^{*(L-2)} \;\;\;\;\;\;\; \text{when} \;N_i \rightarrow \infty,\;\;\;i \in [L-1].
	\]
	We can obtain the convergence value $\mathbf{f}^*$ through the above process. 
	\qed	
}

\bibliography{icml18}
\bibliographystyle{icml2018}

\end{document}